\documentclass{article}

\PassOptionsToPackage{numbers}{natbib}

\usepackage[preprint]{neurips_2021}

\usepackage[utf8]{inputenc} % allow utf-8 input
\usepackage[T1]{fontenc}    % use 8-bit T1 fonts
\usepackage{hyperref}       % hyperlinks
\usepackage{url}            % simple URL typesetting
\usepackage{booktabs}       % professional-quality tables
\usepackage{amsfonts}       % blackboard math symbols
\usepackage{nicefrac}       % compact symbols for 1/2, etc.
\usepackage{microtype}      % microtypography
\usepackage{xcolor}         % colors
\usepackage{amsmath}
\usepackage{graphicx}
\usepackage{subcaption}
\usepackage{multirow}

\title{Exact and soft boundary conditions in\\
       Physics-Informed Neural Networks for the\\
       Variable Coefficient Poisson equation}

\author{%
  Sebastian Barschkis\\
  University of California, Irvine\\
  \texttt{sebbas@sebbas.org} \\
}

\begin{document}

\maketitle

\begin{abstract}
  Boundary conditions (BCs) are a key component in every Physics-Informed Neural Network (PINN)~\cite{raissi2019physics}. By defining the solution to partial differential equations (PDEs) along domain boundaries, BCs constrain the underlying boundary value problem (BVP) that a PINN tries to approximate. Without them, unique PDE solutions may not exist and finding approximations with PINNs would be a challenging, if not impossible task. 
  This study examines how soft loss-based~\cite{raissi2019physics} and exact distance function-based~\cite{sukumar2022exact} BC imposition approaches differ when applied in PINNs. The well known variable coefficient Poisson equation serves as the target PDE for all PINN models trained in this work.
  Besides comparing BC imposition approaches, the goal of this work is to also provide resources on how to implement these PINNs in practice. To this end, Keras~\cite{chollet2015keras} models with Tensorflow~\cite{tensorflow2015-whitepaper} backend as well as a Python notebook with code examples and step-by-step explanations on how to build soft/exact BC PINNs are published alongside this review~\cite{barschkis2023pinn}.
\end{abstract}

\section{Introduction}

Solving the Poisson equation is the most computationally expensive step in many incompressible flow simulations. MFiX's~\cite{syamlal1993mfix} fluid solver, for example, spends more than 80\% of its time on solving the Poisson equation. This finding alone motivates the development of and search for new solvers that can accelerate this step. The advantages of faster Poisson solvers would be manifold: For one, PDE solutions could be found using fewer computational resources. For another, more time could be spent testing and evaluating algorithms that are currently constrained by time spent on solving the Poisson equation.

To this end, we examine PINNs, a promising class of neural PDE approximators. Their performance, training time, and accuracy varies depending on the availability of data but also hinges on how this data is processed. The way a PINN learns labelled data points directly affects training and inference times. To better understand which PINN implementations perform best, this work compares two approaches that can be used to impose boundary conditions: the soft and the exact BC imposition approach. 

These BC imposition methods have been discussed in detail by ~\citet{raissi2019physics} and~\citet{sukumar2022exact}. The primary goal of this work is to show how to embed them in PINNs and provide a flexible, re-usable workflow that can approximate the solution to a PDE. The target PDE in this work is the variable coefficient Poisson equation with Dirichlet boundary conditions.

\section{Related works}

A common approach to learn BCs in PINNs is the soft loss-based approach where a BC data loss is part of the total loss. The deep learning framework that introduced PINNs~\cite{raissi2019physics} followed this approach by incorporating the Mean-squared-error (MSE) of points found along domain boundaries to the overall loss. Using this approach~\citet{raissi2019physics} show that PINNs can successfully learn periodic and Dirichlet boundary conditions while approximating solutions to instances of Schr\"odingers, Burgers, and Navier-Stokes equations. Many other PDEs have been solved with PINNs and similar MSE losses for boundary conditions: \citet{bischof2021multi}, for instance, showed solutions to Kirchhoff's plate bending problem and the Helmholtz equation.

While having exact values at domain boundaries constrains the problem space, BCs are a sufficient and not a necessary condition for PINNs to converge. If knowledge on domain boundaries is given in another form, PINNs can still find PDE solutions. \citet{raissi2020hidden} trained PINNs that successfully approximate fluid velocities and pressure in laminar flows even in the absence of specific boundary conditions. They show that a concentration field for a passive scalar with sufficient gradients near domain boundaries can compensate the lack of explicit BCs.

Recent works have looked into hard constraints for PDEs in PINNs. \citet{negiar2022learning} developed a differentiable PDE-constrained layer (PDE-CL) through which hard constraints can be placed on the interior of a domain. Their approach stands in contrast to previous work with hard constraints which finds PDE solutions by enforcing hard constraints at domain boundaries~\citet{lu2021physics, sukumar2022exact}.

The comparisons shown in this work will make use of the commonly used soft BC imposition approach introduced by~\citet{raissi2019physics}. For exact BCs, \citet{sukumar2022exact}'s distance-based method will be used.

\section{Background}

All PINNs from this study were trained to approximate Poisson's equation with Dirichlet boundary conditions. Special cases of Poisson's equation, such as Laplace's equation and Poisson's equation with zero-BC, were examined as well.

\subsection{Poisson's equation}

We consider Poisson's equation in a 2-dimensional (2D) setting. More precisely, the PINNs from this work use the variation of Poisson's equation with variable coefficient and Dirichlet boundary conditions which can be expressed as follows

\begin{align}
    \nabla \cdot (a \nabla p(x,y)) &= f(x,y), \qquad (x,y) \in \Omega \label{eq:poisson}\\
    p(x,y) &= g(x,y), \qquad (x,y) \in \partial\Omega \label{eq:dirichlet_bc}
\end{align}

where $a$ is the variable coefficient, $f$ the right-hand side (RHS), $g$ the boundary condition (BC), and $p$ the solution.

\subsection{Variations of Poisson's equation}

By forcing certain variables in Poisson's equation~\eqref{eq:poisson} to zero we obtain PDE variations that can easily be approximated with the same PINNs. This work considers the cases where either $f=0$ or $g=0$. That is, Laplace's equation in the former and the zero BC Poisson equation in the latter case.

\subsection{Data generation}\label{ch:data_gen}

\begin{figure}[ht]
    \centering
    \subcaptionbox{XY-coordinates and knots\label{subfig:knots}}{\includegraphics[height=.3\textwidth]{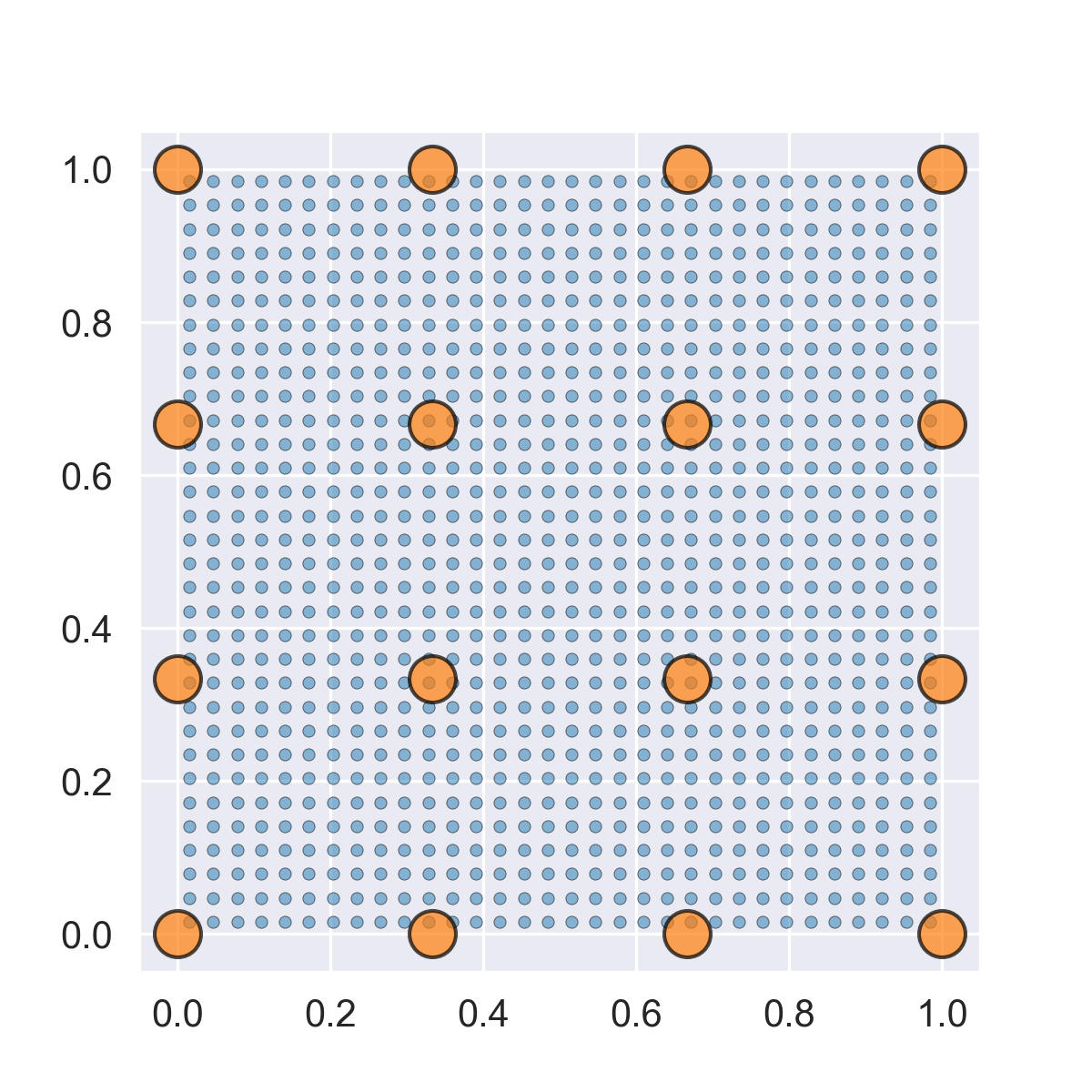}}%
    \subcaptionbox{Example for $a$\label{subfig:a}}{\includegraphics[height=.28\textwidth]{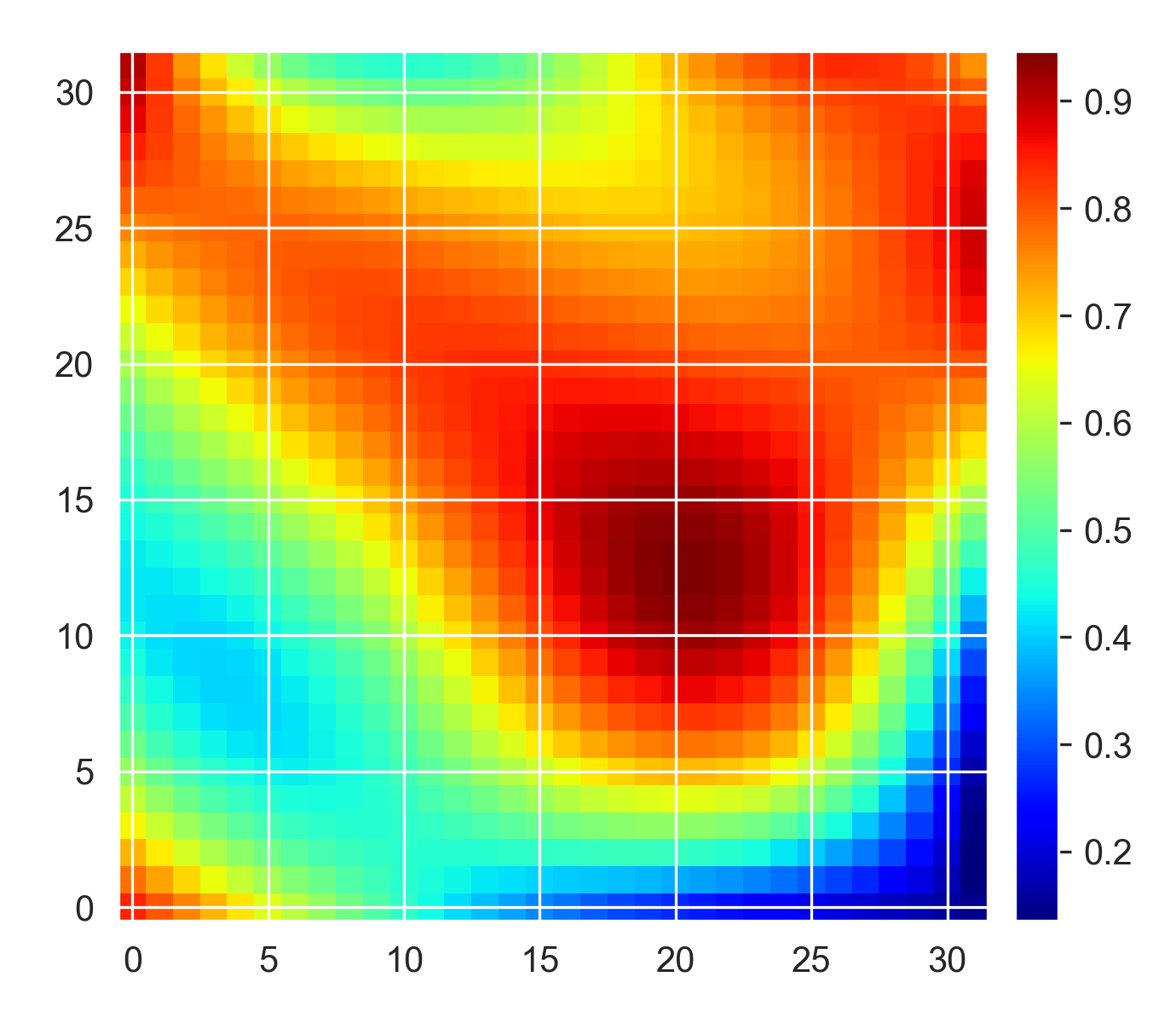}}%
    \subcaptionbox{Example for $f$\label{subfig:f}}{\includegraphics[height=.28\textwidth]{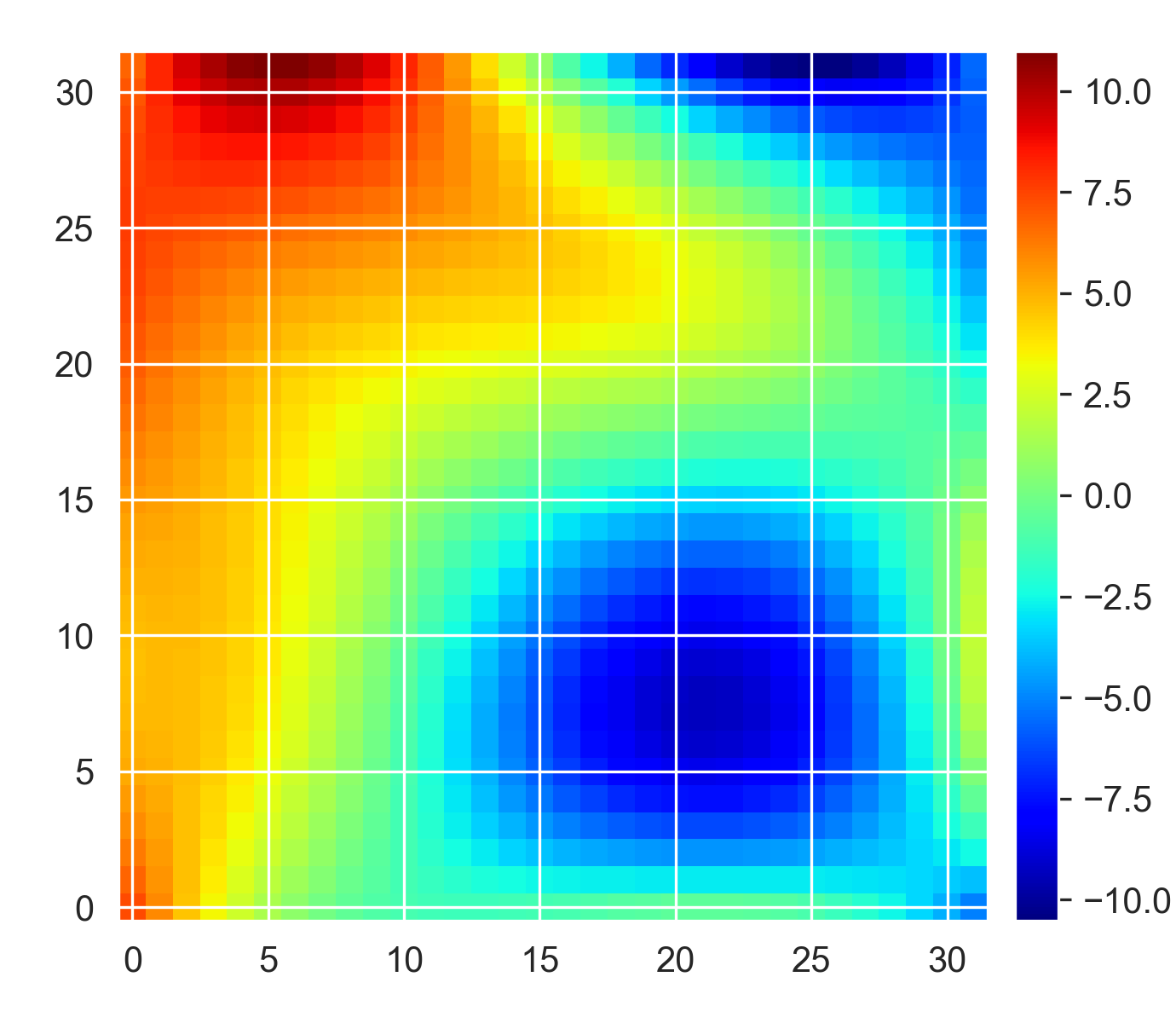}}%
    \caption{Knots (4x4) in domain and interpolated instances for $a$ and $f$ (32x32). \eqref{subfig:knots}: Knots (big dots) and x-y coordinates (small dots) in domain space. \eqref{subfig:a}: $a$ after interpolating from knots. \eqref{subfig:f}: $f$ after interpolating from knots.}
\end{figure}

Training and validation data for coefficient $a$, RHS $f$, and BC $g$ were generated artificially (i.e. not derived from experiments), uniformly and randomly. The generation process followed the following strategy:
\begin{enumerate}
    \item Distribute $n \times n$ knots uniformly and in a grid pattern on the  discretized target domain (Figure~\ref{subfig:knots}).
    \item Generate a Sobol sequence with given minimum and maximum values.
    \item For each knot, select a value from the Sobol sequence.
    \item Use a Gaussian process to interpolate from the values found at knots and sample interpolated values at every discrete x-y domain position.
    \item Repeat this process for variables $a$, $f$, and $g$.
\end{enumerate}

Note that for $g$, the above process was executed in 1-dimensional (1D) setting since for BCs only a vector that wraps around the 2D domain is needed.
This data generation approach ensures that the resulting value distributions in $a$, $f$, and $g$ are smooth and not completely random. I.e. the resulting datasets are free of discontinuities. Since PINN models compute physics losses and expect training data to be differentiable, smooth value distributions are highly desirable.

\section{Soft and exact BCs in PINNs}

\subsection{Loss function}
The total loss consists of a supervised data and a PDE loss:
\begin{align*}
    \mathcal{L}_{Total} = \mathcal{L}_{Data} + \alpha \cdot \mathcal{L}_{PDE}
\end{align*}
Both $\mathcal{L}_{Data}$ and $\mathcal{L}_{PDE}$ use the Mean-squared-error (MSE) metric.

$\mathcal{L}_{Data}$ is used to compute the error for points near the domain boundary. While soft BC models rely on minimized errors at these points, exact BC models are less impacted by them as they enforce BCs at domain boundaries and, hence, errors in these areas will be very small anyways. 
$\mathcal{L}_{PDE}$ on the other hand is plays a key role in both soft and exact BC models. It measures the error between $\nabla\cdot (a \nabla p)$ and RHS $f$. The solution in the domain interior depends on this error.

\subsection{Soft BC models}

\begin{figure}[ht]
    \centering
    \subcaptionbox{Collocation points (sampled uniformly and randomly)\label{subfig:colloc_pts}}{\includegraphics[width=.35\textwidth]{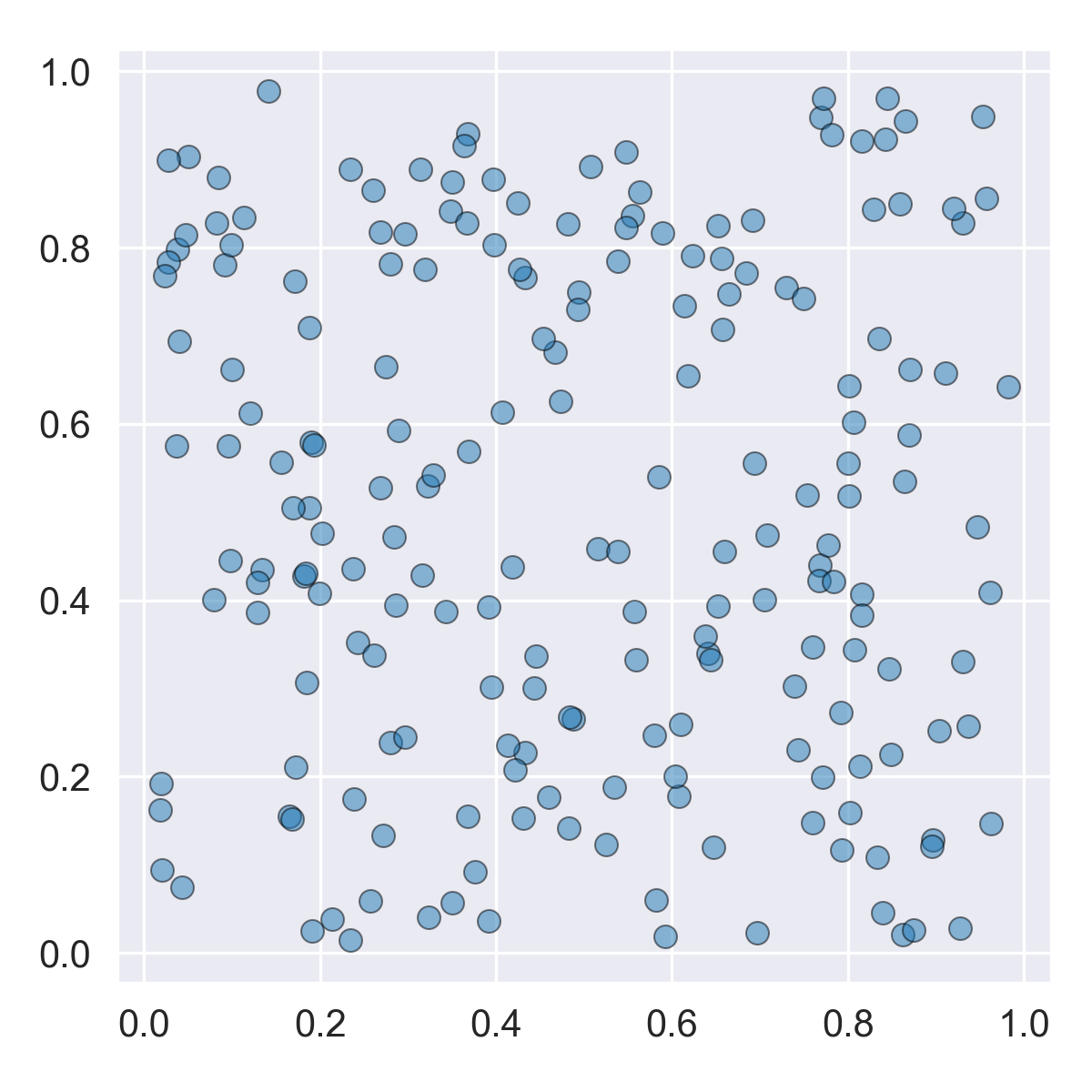}}\hspace{3em}%
    \subcaptionbox{Same collocation points plus boundary points (sampled uniformly)\label{subfig:colloc_bnd_pts}}{\includegraphics[width=.35\textwidth]{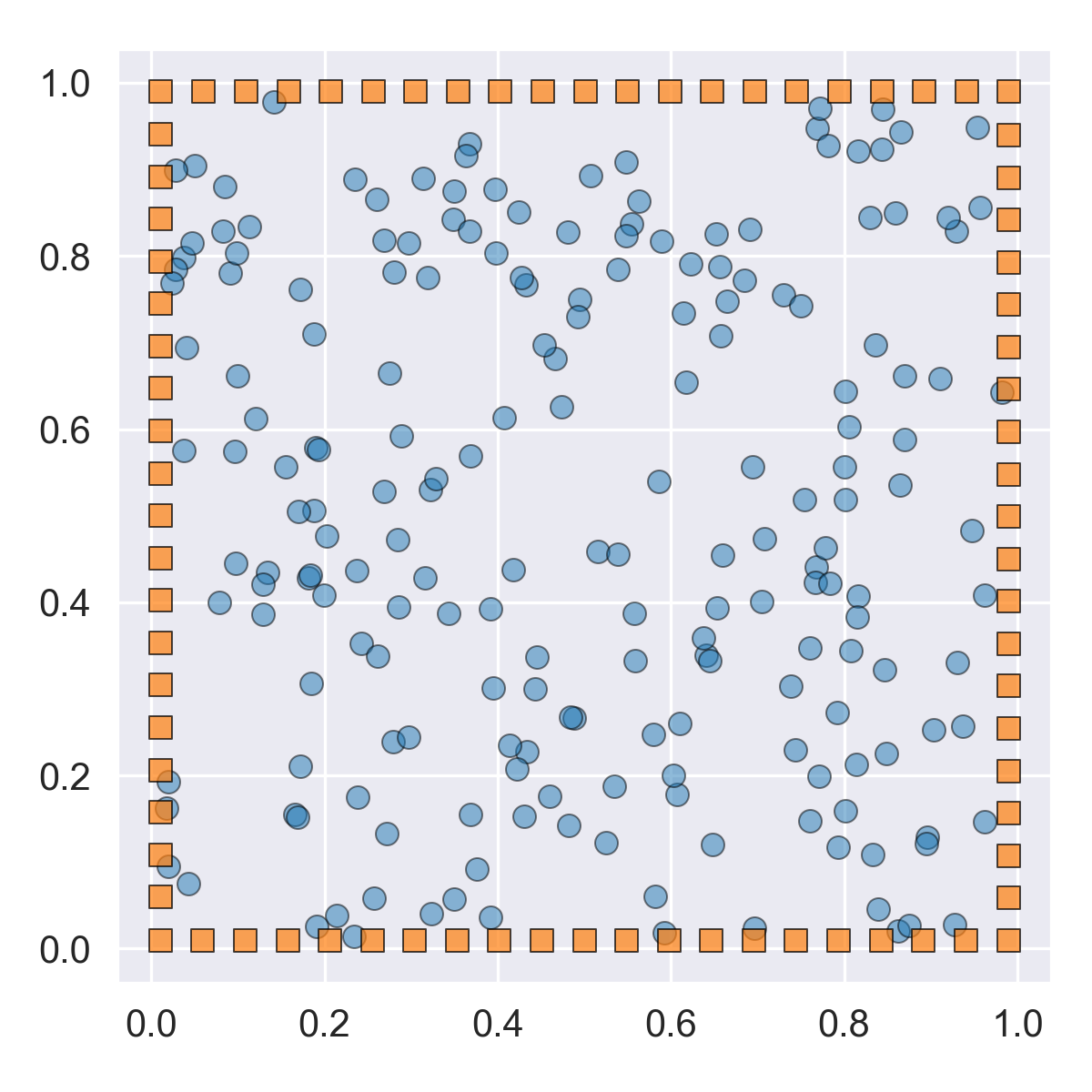}}%
    \caption{Points sampled in domain space: \eqref{subfig:colloc_pts} 200 collocation points, sampled randomly and uniformly. \eqref{subfig:colloc_bnd_pts} Collocation points with 20 additional boundary points per domain side. (Note: number of points shown only for illustration purposes, not used in actual models).}
\end{figure}

When sampling points uniformly and randomly in domain space, the number of points placed at domain boundaries is usually insufficient for PINNs to learn BCs efficiently. As a result, training becomes inefficient since the PDE loss computed at collocation points found in the domain interior cannot converge sufficiently due to the lack of learned BC knowledge. A common strategy that compensates this behavior is to selectively sample more points near domain boundaries (Figure ~\ref{subfig:colloc_bnd_pts}). These additional domain boundary points make it possible to minimize the data loss $\mathcal{L}_{Data}$ quicker and consequently also help to find the PDE solution values that minimize $\mathcal{L}_{PDE}$ in the domain interior.

\subsection{Exact BC models}

\begin{figure}[ht]
    \centering
    \includegraphics[width=.75\textwidth]{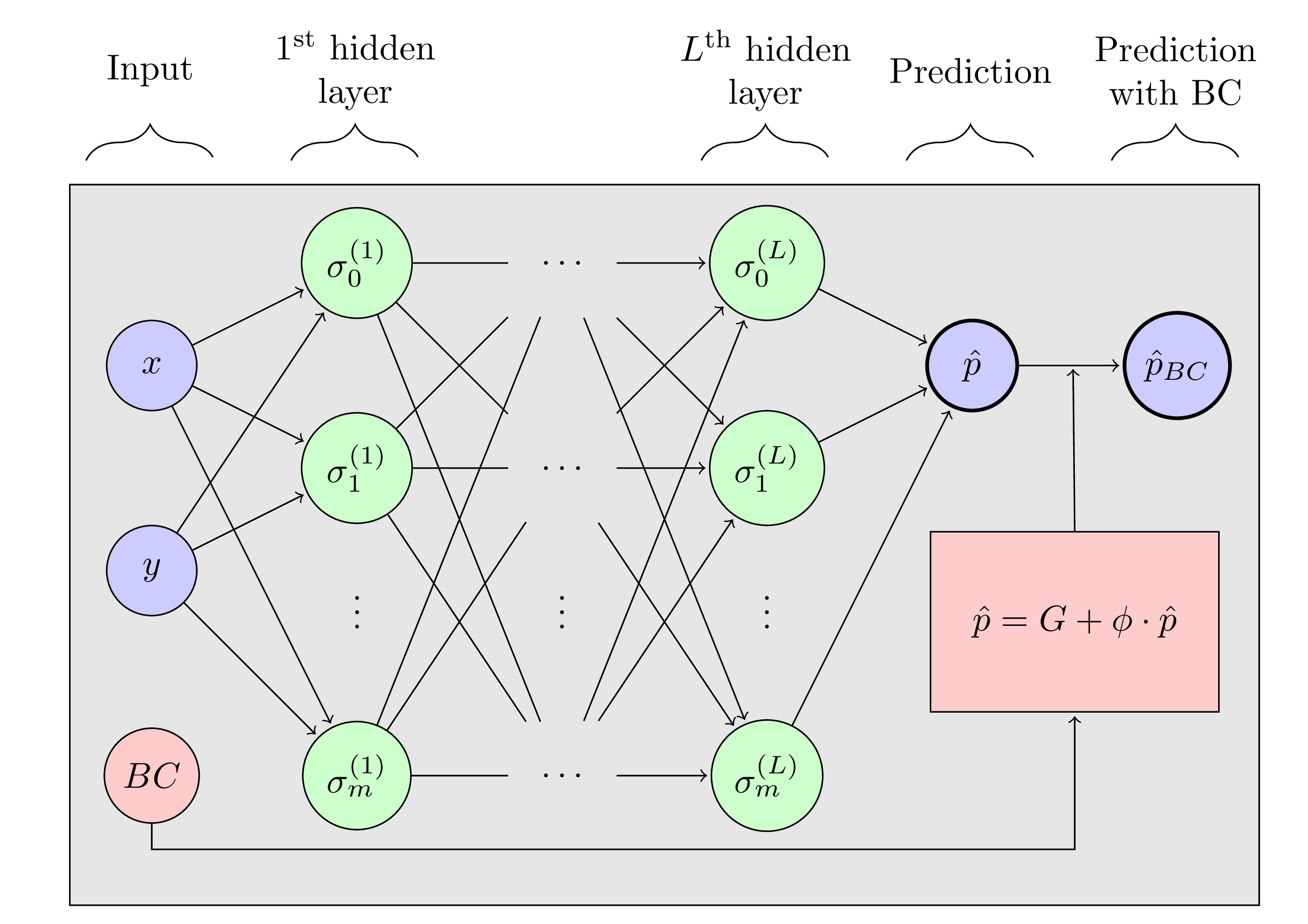}
    \caption{Exact BC enforcement in a fully-connected architecture: The BC is ``injected'' into the result tensor.}
    \label{fig:overview_exactbc}
\end{figure}

The exact BC imposition approach with distance functions follows work from~\citet{sukumar2022exact}. Their approach is based on the idea that instead of learning BCs through a loss term, BCs can be hard-coded into PINNs too. By reformulating network solutions as functions of an interpolation and a filter function, the exact BC becomes part of the predicted solution. I.e. the approximation $\hat{p}$ from a neural network with exact BCs can be expressed as

\begin{align*}
    \hat{p}(x,y) = G(x, y) + \phi(x, y) \cdot \hat{p}(x,y)
\end{align*}

where $G(x,y)$ is the interpolating and $\phi(x,y)$ the filter function.

\subsubsection{Interpolation function}

$G(x,y)$ is the function that interpolates BCs. That is, at any position $(x,y)$ in the domain $G$ returns an approximated solution value that is only based on the values found at domain boundaries. By using an interpolation method such as Inverse-Distance-Weighting (IDW), it is possible to broadcast the BC to the domain interior with varying weight. IDW achieves this by measuring the distance from collocation points to each boundary point and then computing the weighted solution value based on the distance to the BC points.

\begin{align*}
G(\boldsymbol{x}) &= 
\sum_{i}^{N_{bc}}
  \frac{
    w_{i}z_{i}
  }{
    \sum_{i}^{N_{bc}}{
      w_{i}
    }
  }\\
w_i &= \lvert\boldsymbol{x}-\boldsymbol{x_i}\rvert
\end{align*}

Note that special treatment is required in situations where a collocation point coincides with a BC point, i.e. the case where $w_i=0$. The resulting zero-division can be avoid in practice by adding $\epsilon>0$ to $w_i$.

\subsubsection{Filter function}

Filter function $\phi(x,y)$ forms the second part in exact BC models. It scales the network and reduces output contributions in areas close to the BC. $\phi(x,y)$ should be smooth and chosen with domain knowledge in mind. Depending on the shape of the domain, the domain sides where exact BCs should be enforced, and the overall filtering strength, $\phi(x,y)$ can take different shapes. For square domains and exact BCs along all domain sides, a good choice is

\begin{align*}
    \phi(x,y) = x \cdot (1-x) \cdot y \cdot (1-y)
\end{align*}

where $x$ and $y$ are locations in a domain with width and height in range $[0,1]$.

\section{Experiments}

The PINN models with soft and exact BCs were trained on instances of the Poisson equation with non-zero BC, the Laplace equation, and the Poisson equation with zero-BC. To ensure comparability, all datasets had the same dimension (128x128) and were used to train one soft BC and one exact BC model. I.e. every side-by-side comparision of soft and exact BC models is the result from training with the same instances of $a$, $f$, and $g$.

All models were trained using fully-connected architectures with 4 layers and 128 neurons per layer. $GELU$ activations were used in all layers except for the last layer which used a linear activation. The optimizer was a Adam optimizer with $\beta_1=0.9$, $\beta_2=0.999$ and initial learning rate of $0.0005$. Learning rate was dropped on plateau with factor $0.1$ after a patience of $10$ epochs and $\delta_{min} = 0.01$.

In both soft and exact BC models, the loss was computed using $10.000$ (collocation) points, each of which contributed to the PDE or data loss. The latter was only computed for points in vicinity to domain boundaries. In soft BC models, an additional $1.000$ boundary points were sampled near domain boundaries. Exact BC models did not make use of additional boundary points. Instead, to compute $G$, they used $20$ boundary condition points per domain side.

\subsection{Parameter selection}

$GELU$ was chosen over $tanh$ as the activation function since models tended to yield slightly lowers errors in the former case. The choice, however, is highly dependent on training data. Poisson datasets were the random generator produced narrow value distributions (i.e. datasets were values at knots (see section~\ref{ch:data_gen}) were close to each other) benefited more from $tanh$ activations.

The choice of $\alpha=0.1$ was based on the assumption that the number of collocation and boundary points has a ratio of 10:1. Exact BC models don't have to learn the BC through a loss term and could use a larger $\alpha$. However, to ensure fairness during training, $\alpha=0.1$ was used in both soft and exact BC models.

\subsection{Poisson's equation with non-zero BC}\label{sec:poisson_nonzero_exp}

\begin{figure}[ht]
\captionsetup[subfigure]{labelformat=empty,justification=centering}
\centering
\subcaptionbox{Ground Truth}{\includegraphics[height=.184\textwidth]{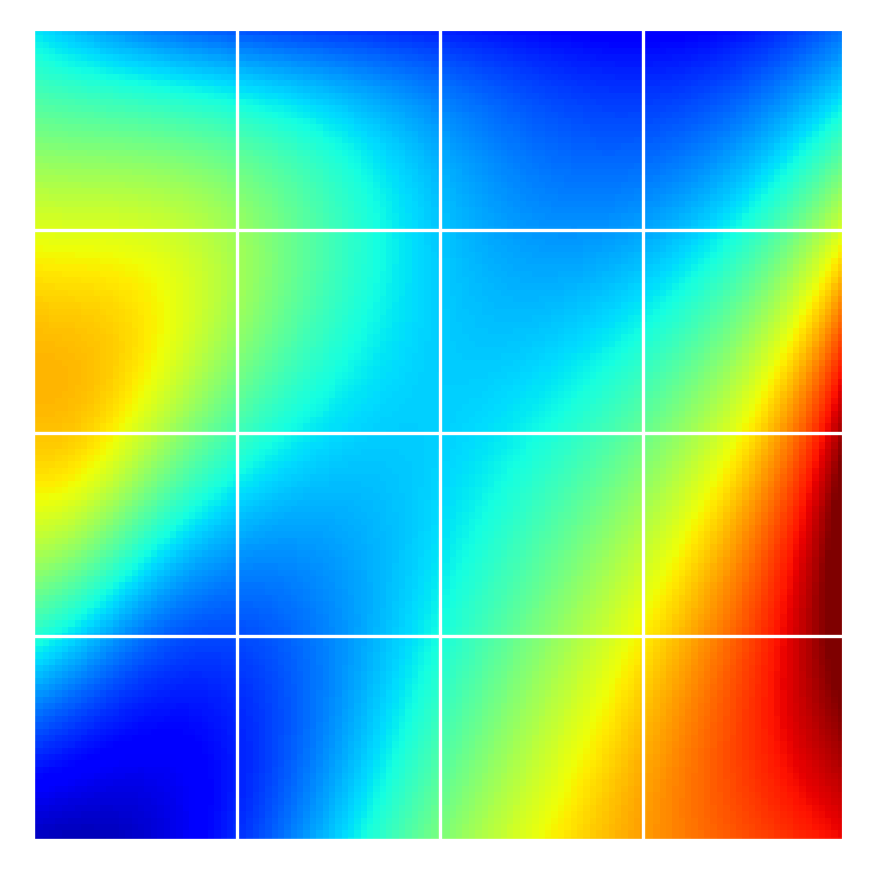}}%
\subcaptionbox{Prediction\\Soft BC}{\includegraphics[height=.184\textwidth]{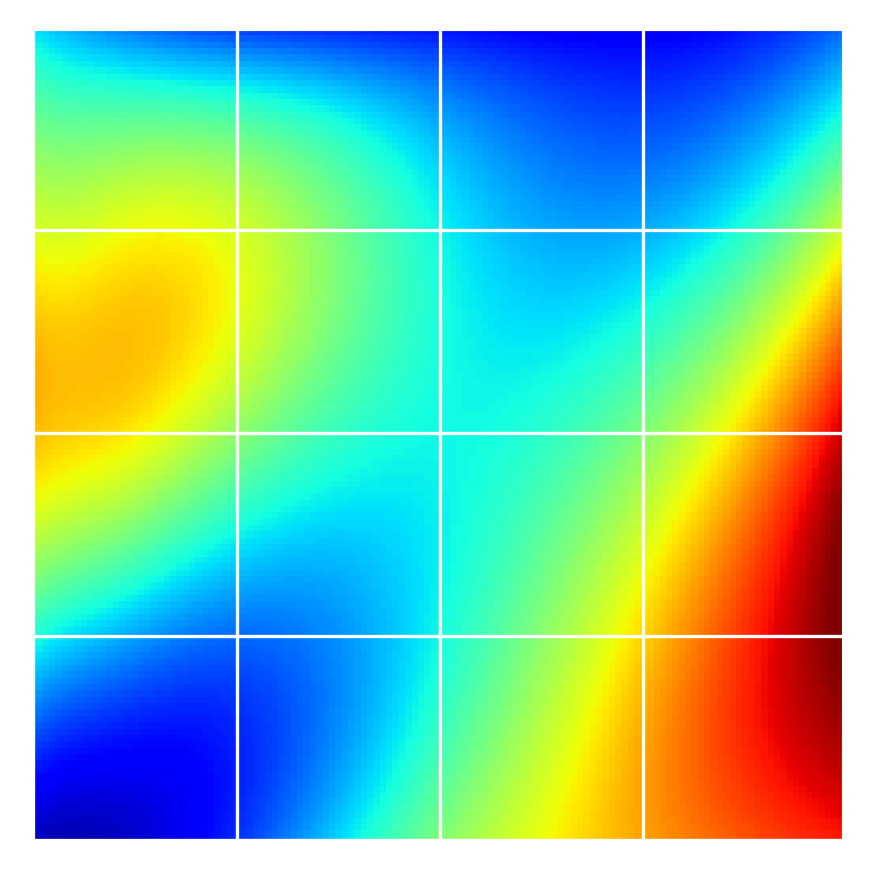}}%
\subcaptionbox{Prediction\\Exact BC}{\includegraphics[height=.184\textwidth]{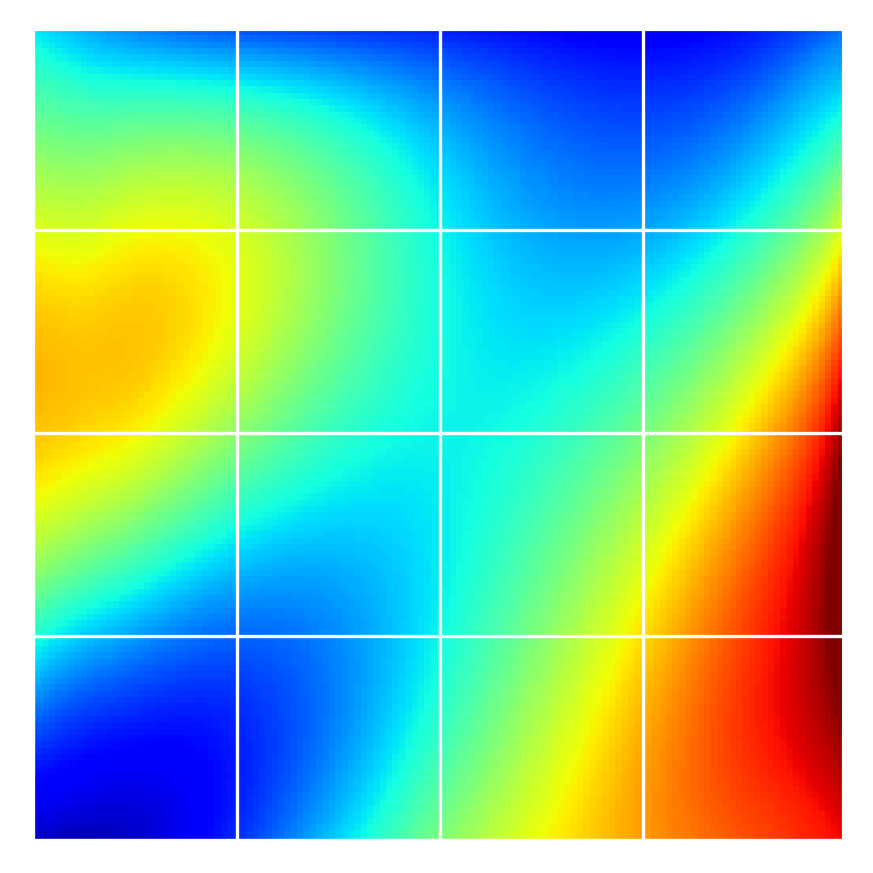}}%
\subcaptionbox{}{\includegraphics[height=.184\textwidth]{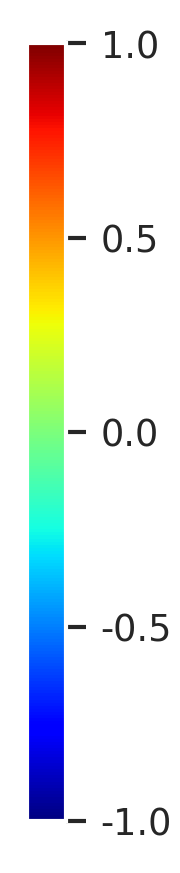}}%
\subcaptionbox{Error\\Soft BC}{\includegraphics[height=.184\textwidth]{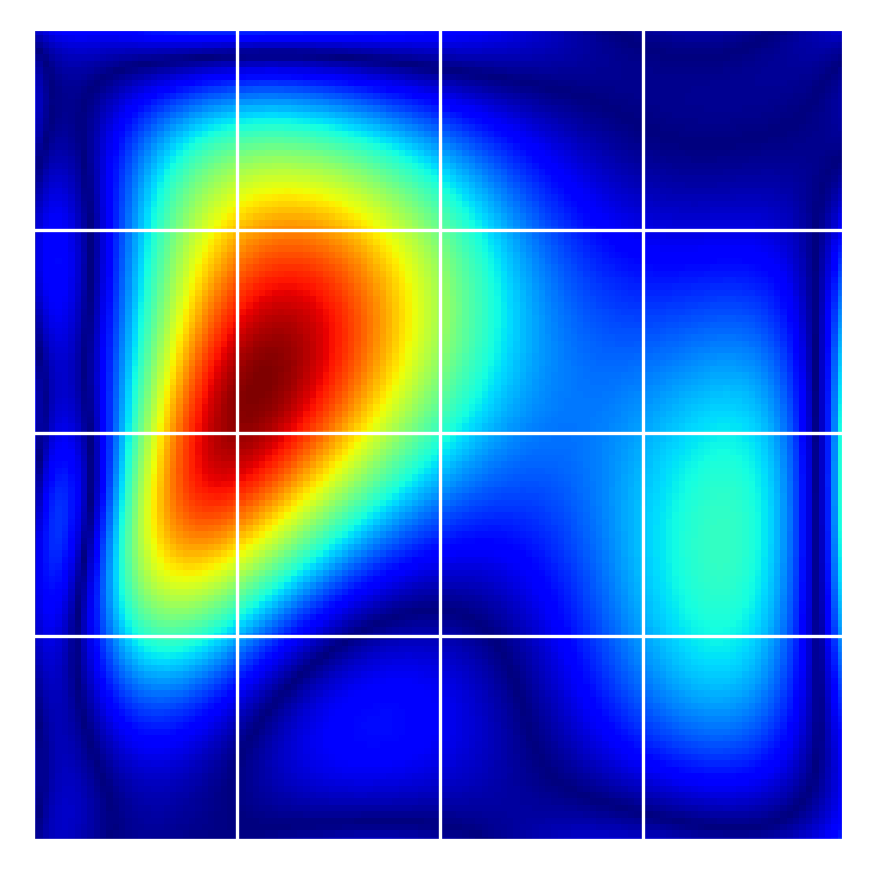}}%
\subcaptionbox{Error\\Exact BC}{\includegraphics[height=.184\textwidth]{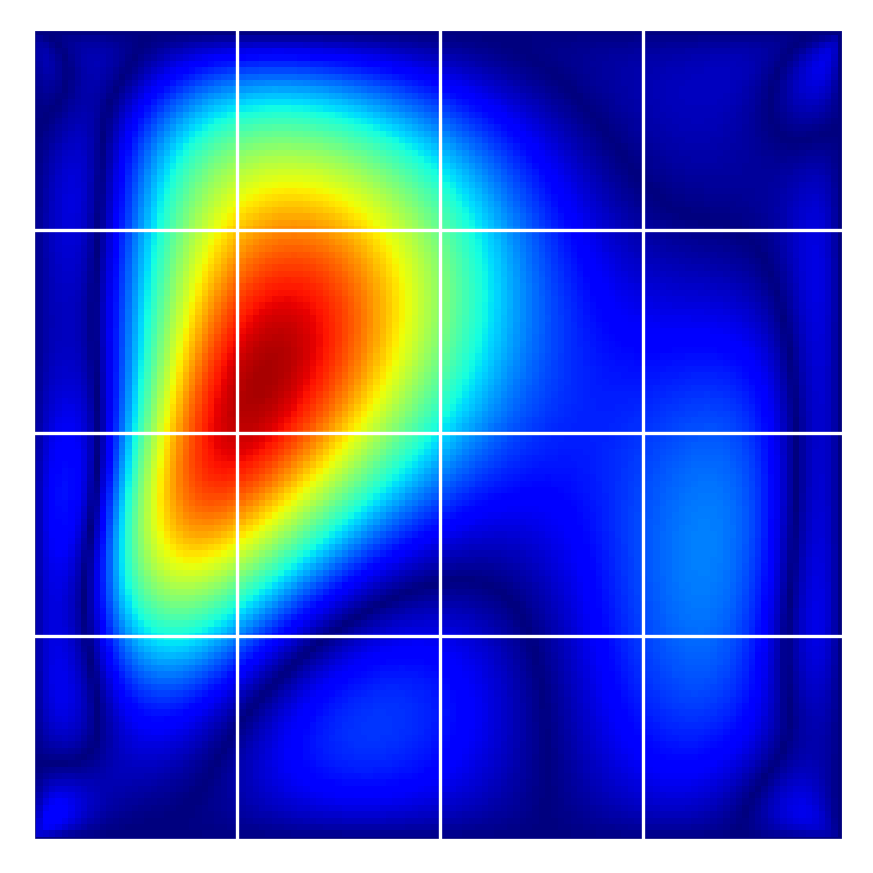}}%
\subcaptionbox{}{\includegraphics[height=.184\textwidth]{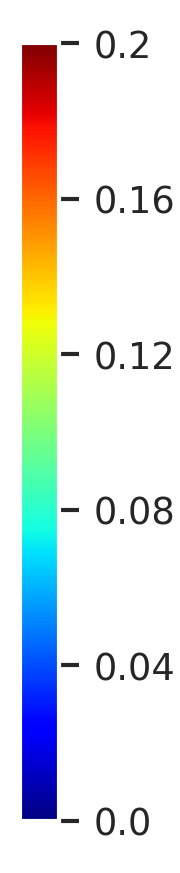}}
\caption{Results from models solving Poisson's equation with non-zero BC.}
\label{fig:models_eq2_case2}
\end{figure}

The plots in Figure~\ref{fig:models_eq2_case2} show the results for solution $p$ to an instance of Poisson's equation with non-zero BC. Grids for $a$ and $g$ resulted from knots with random values in range $[-1, 1]$. $f$ was generated with knots ranging from $[-10,10]$.

\begin{table}[ht]
  \centering
  \begin{tabular}{clcccc}
    \toprule
    Dataset \#         & BC Type & MAE     & RMSE    & MAPE      & Avg. time / epoch (sec)\\
    \midrule
    \multirow{2}{*}{0} & Soft    & 4.97e-2 & 7.01e-2 & 114.55\%  & 1.78         \\
                       & Exact   & 4.29e-2 & 6.46e-2 & 116.48\%  & 1.86         \\
    \bottomrule\\
  \end{tabular}
  \caption{Error metrics for models solving Poisson's equation with non-zero BC (plots in Figure~\ref{fig:models_eq2_case2})}
  \label{tab:models_eq2_case2}
\end{table}

\subsection{Laplace's equation}\label{sec:laplace_exp}

\begin{figure}[ht]
\captionsetup[subfigure]{labelformat=empty,justification=centering}
\centering
\subcaptionbox{Ground Truth}{\includegraphics[height=.184\textwidth]{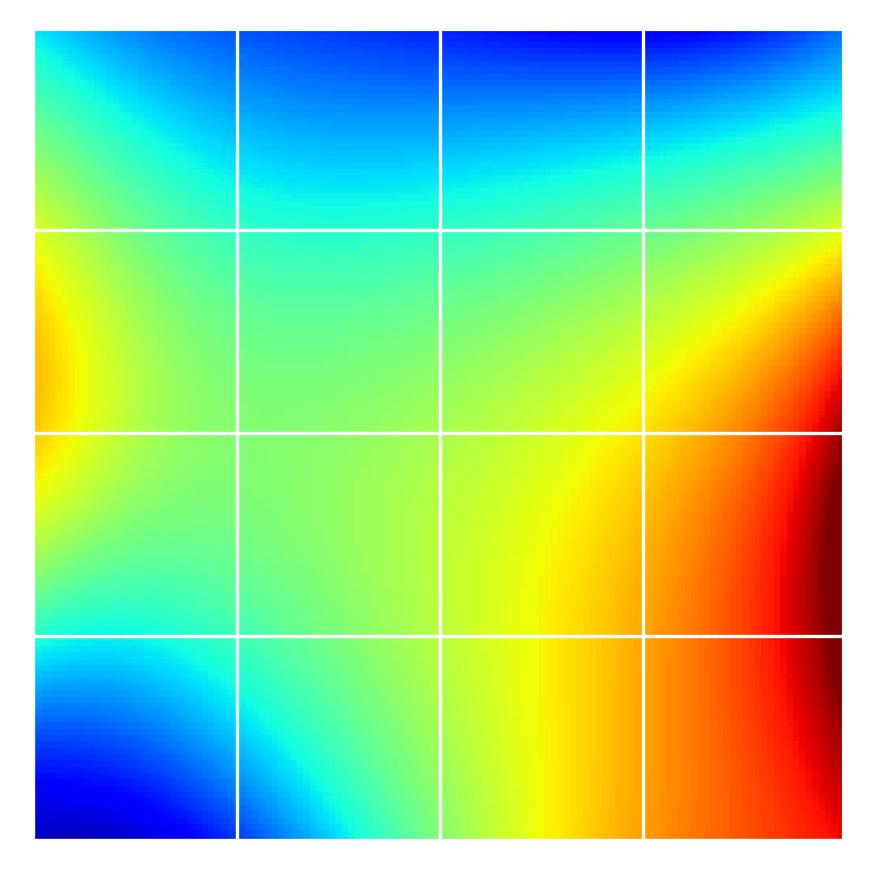}}%
\subcaptionbox{Prediction\\Soft BC}{\includegraphics[height=.184\textwidth]{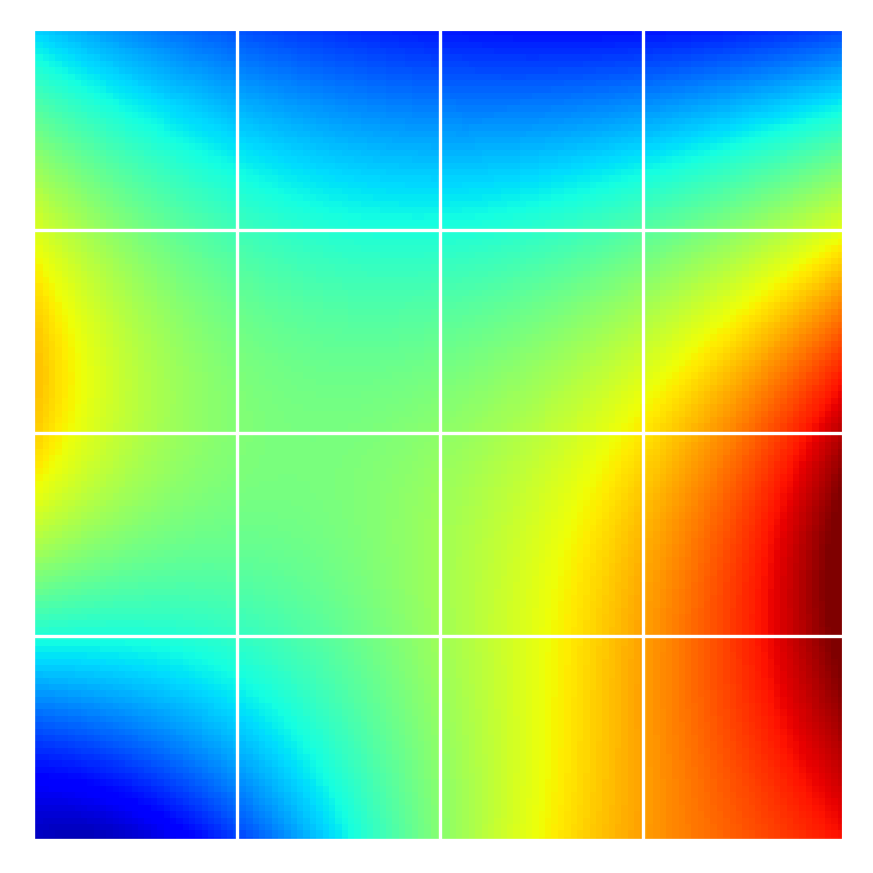}}%
\subcaptionbox{Prediction\\Exact BC}{\includegraphics[height=.184\textwidth]{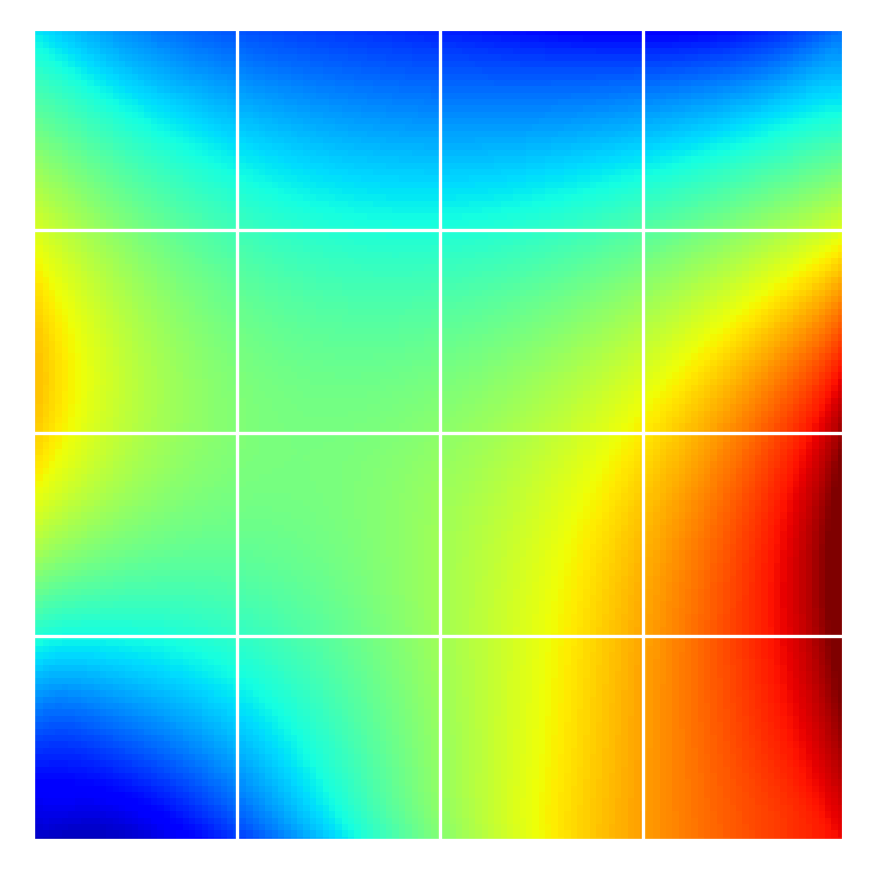}}%
\subcaptionbox{}{\includegraphics[height=.184\textwidth]{figs/colorbar/colorbar_1-1.png}}%
\subcaptionbox{Error\\Soft BC}{\includegraphics[height=.184\textwidth]{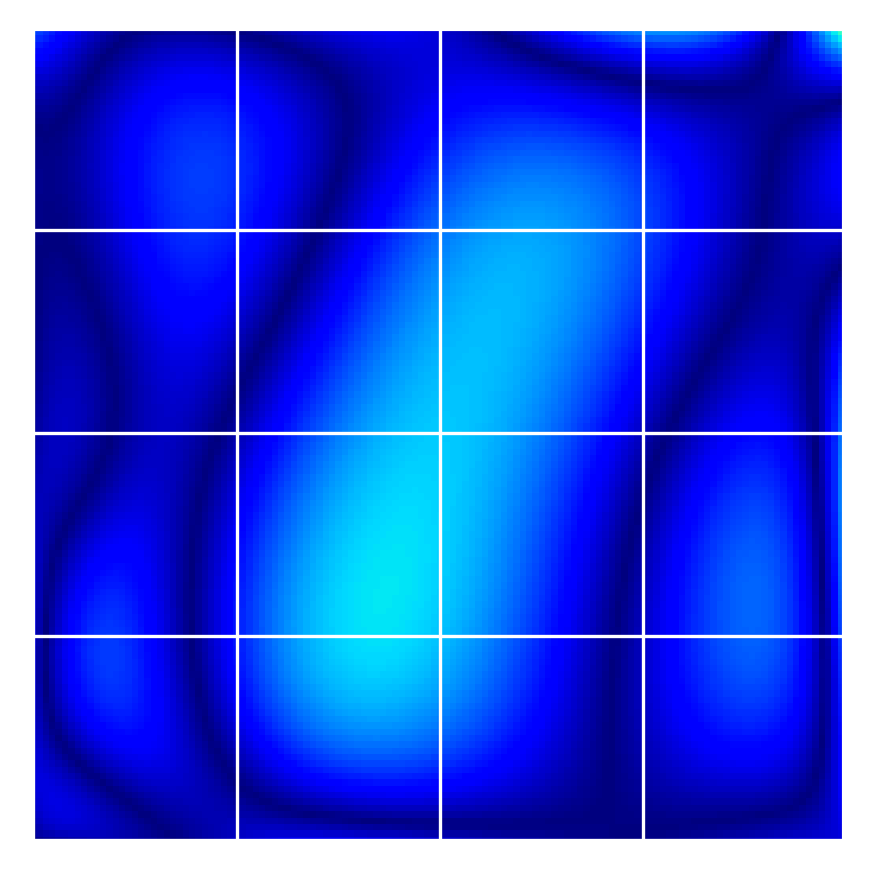}}%
\subcaptionbox{Error\\Exact BC}{\includegraphics[height=.184\textwidth]{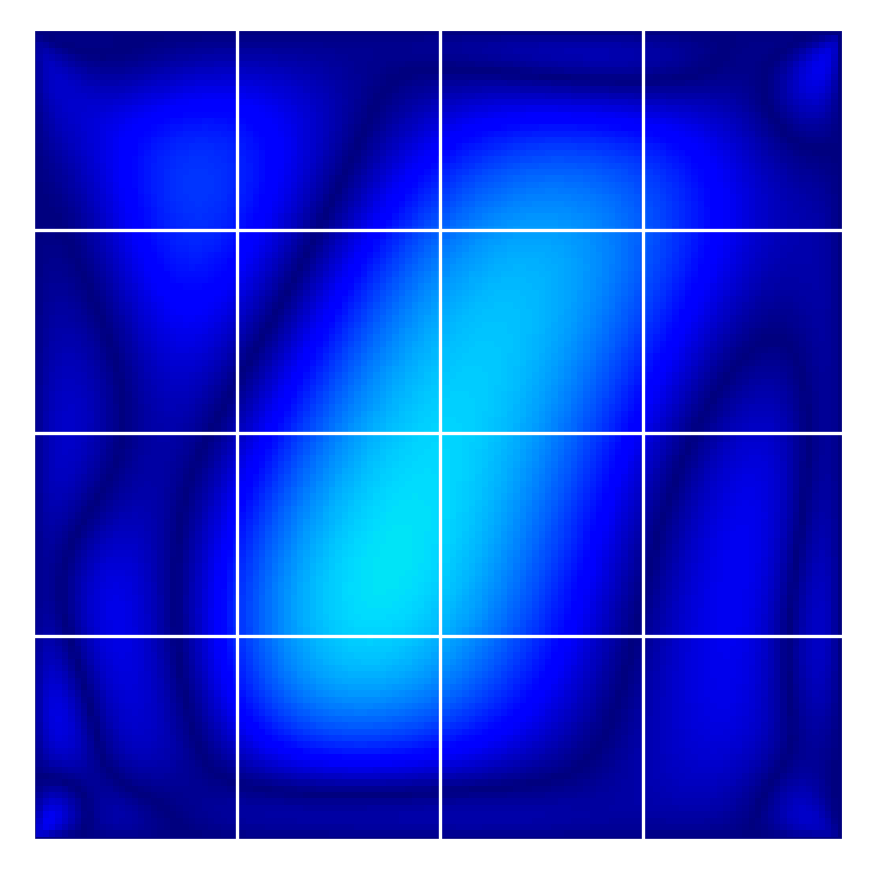}}%
\subcaptionbox{}{\includegraphics[height=.184\textwidth]{figs/colorbar/colorbar_02-0.png}}
\caption{Results from models solving Laplace's equation with non-zero BC.}
\label{fig:models_eq2_case1}
\end{figure}

The models approximating Laplace's equation, shown in Figure~\ref{fig:models_eq2_case1}, are based on the same training data used in the models for Poisson's equation with non-zero BC (Section~\ref{sec:poisson_nonzero_exp}). Only RHS $f=0$ differs.

\begin{table}[!h]
  \centering
  \begin{tabular}{clcccc}
    \toprule
    Dataset \#         & BC Type & MAE     & RMSE    & MAPE      & Avg. time / epoch (sec)\\
    \midrule
    \multirow{2}{*}{0} & Soft    & 2.53e-2 & 3.19e-2 & 150.71\%  & 1.79         \\
                       & Exact   & 2.19e-2 & 2.99e-2 & 140.02\%  & 1.84         \\
    \bottomrule\\
  \end{tabular}
  \caption{Error metrics for models solving Laplace's equation with non-zero BC (plots in Figure~\ref{fig:models_eq2_case1})}
  \label{tab:models_eq2_case1}
\end{table}

\subsection{Poisson's equation with zero BC}\label{sec:poisson_zero_exp}

Similarly to the experiments with non-zero BCs, the zero-BC model for Poisson was trained with the same $a$ and $f$. Note how since $g=0$, the solutions $p$ are on a different scale.

\begin{figure}[!h]
\captionsetup[subfigure]{labelformat=empty,justification=centering}
\centering
\subcaptionbox{Ground Truth}{\includegraphics[height=.184\textwidth]{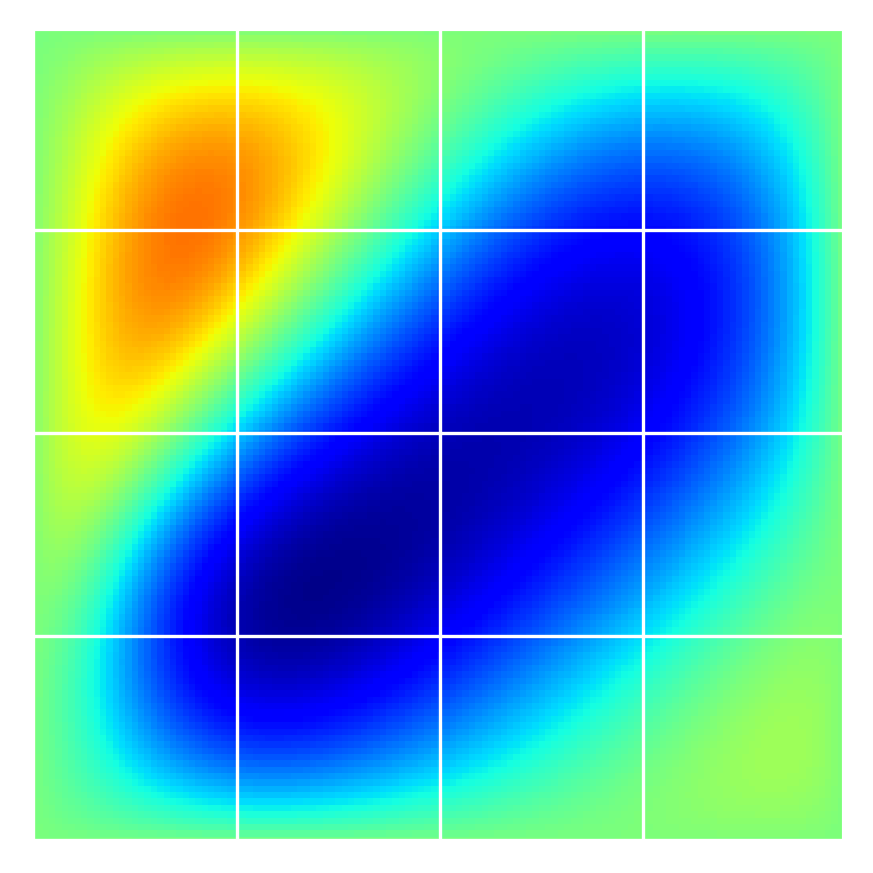}}%
\subcaptionbox{Prediction\\Soft BC}{\includegraphics[height=.184\textwidth]{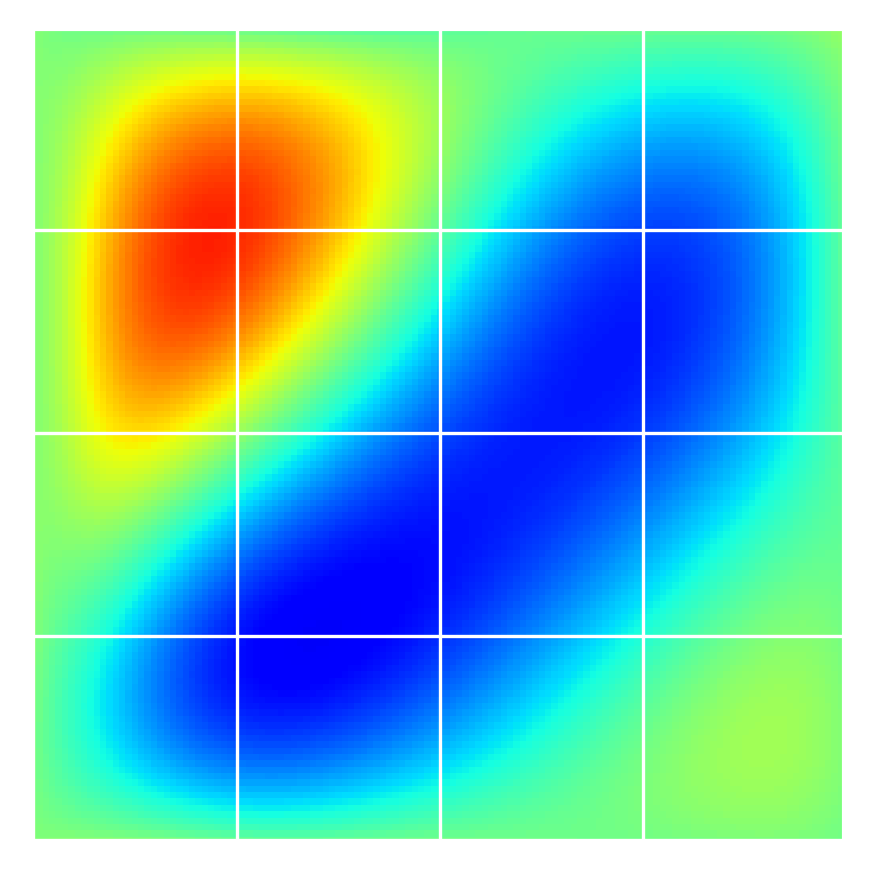}}%
\subcaptionbox{Prediction\\Exact BC}{\includegraphics[height=.184\textwidth]{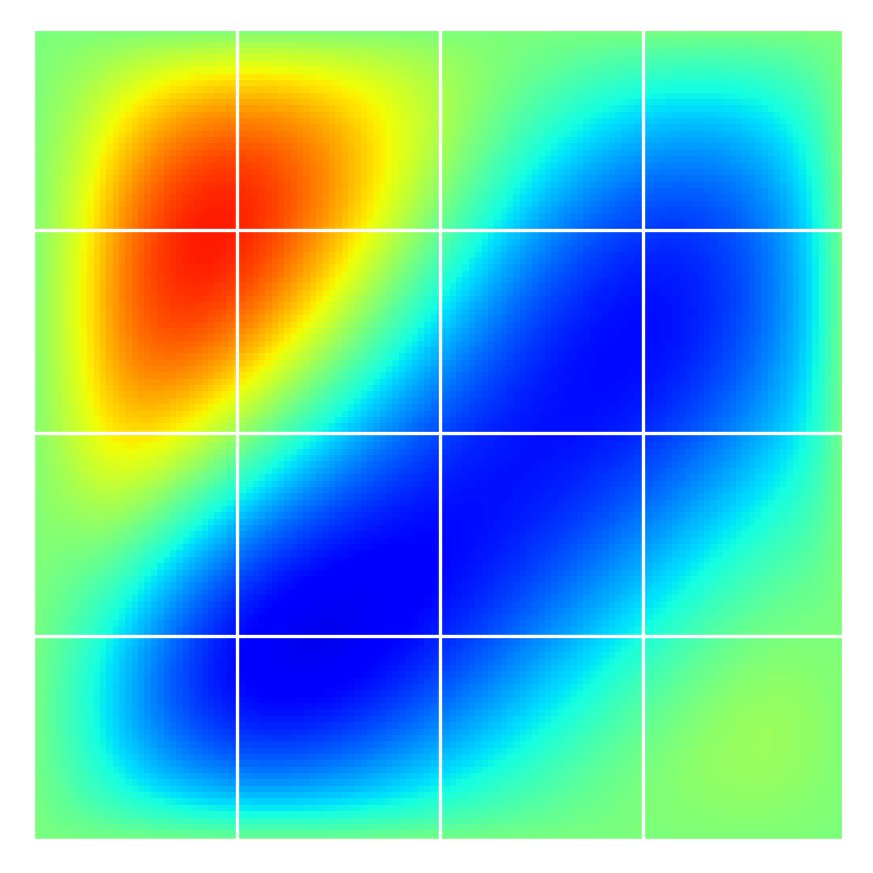}}%
\subcaptionbox{}{\includegraphics[height=.184\textwidth]{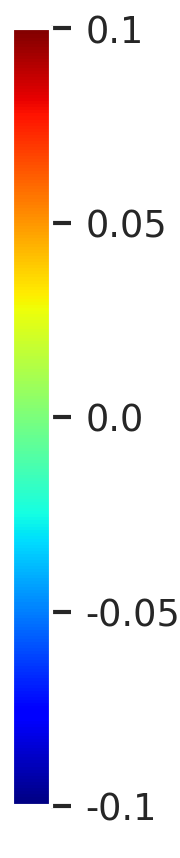}}%
\subcaptionbox{Error\\Soft BC}{\includegraphics[height=.184\textwidth]{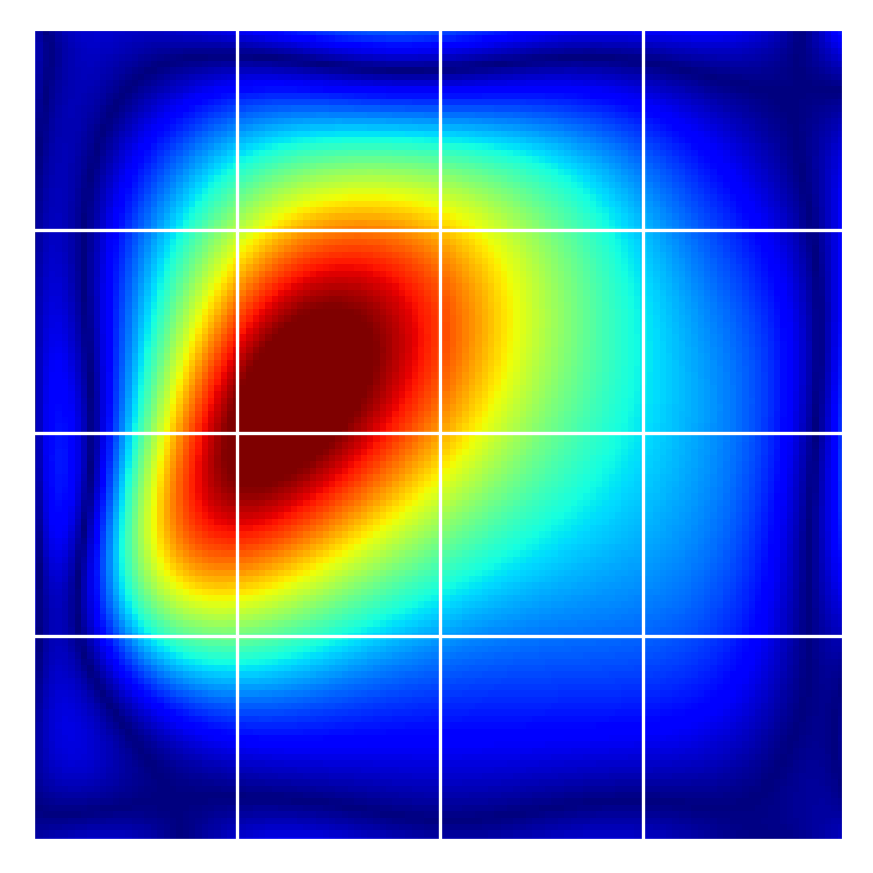}}%
\subcaptionbox{Error\\Exact BC}{\includegraphics[height=.184\textwidth]{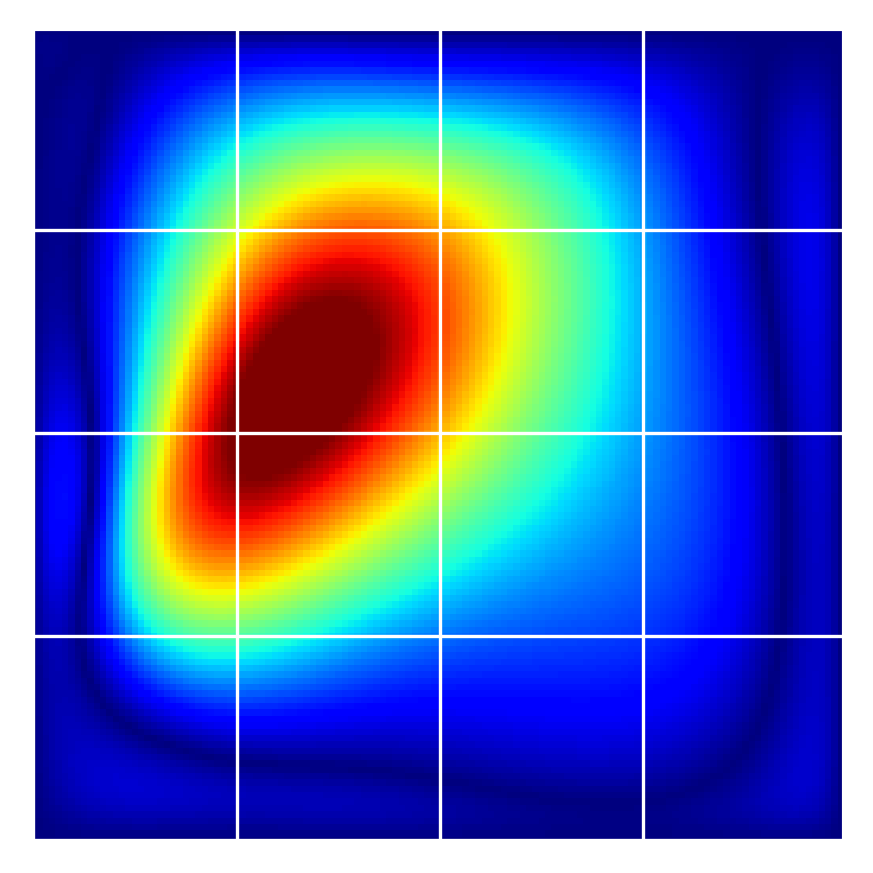}}%
\subcaptionbox{}{\includegraphics[height=.184\textwidth]{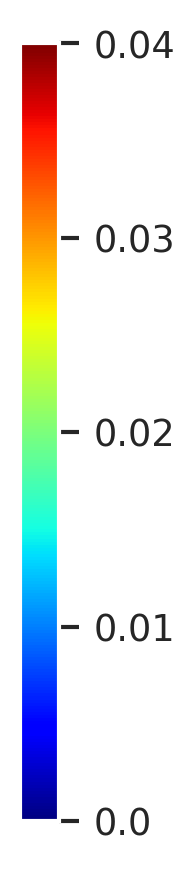}}
\caption{Results from models solving Poisson's equation with non-zero BC.}
\label{fig:models_eq2_case0}
\end{figure}

\begin{table}[!h]
  \centering
  \begin{tabular}{clcccc}
    \toprule
    Dataset \#         & BC Type & MAE     & RMSE    & MAPE      & Time / epoch (sec)\\
    \midrule
    \multirow{2}{*}{0} & Soft    & 6.08e-2 & 8.39e-2 & 152.46\%  & 1.77         \\
                       & Exact   & 5.79e-2 & 8.27e-2 & 119.99\%  & 1.86         \\
    \bottomrule\\
  \end{tabular}
  \caption{Error metrics for models solving Poisson's equation with non-zero BC (plots in Figure~\ref{fig:models_eq2_case0})}
  \label{tab:models_eq2_case0}
\end{table}

\section{Conclusions}

The experiments with both Poisson's and Laplace's equation have shown that exact BC models generally yield smaller errors than soft BC models. Since exact BC models don't have to learn boundary points, training can focus on collocation points and minimizing the PDE loss. I.e. all effort goes into learning areas where no labelled data is present and where hidden physics should be recovered.

However, exact BC models are not always the better choice. The following points should be considered when using exact BCs:

\begin{itemize}
    \item When increasing the number of BC points in exact BC models, both training and inference times will be prolonged due to the interpolation overhead (every collocation point has to evaluate every BC point). 
    \item The number of BC points in exact BC models should be chosen carefully and with the domain size in mind. A small domain will require fewer BC points than a larger domain.
    \item Filter function $\phi$ should be chosen carefully as well. Especially when the domain is non-square or when boundary conditions are not enforced everywhere. An inappropriate choice of $\phi$ can significantly increase training times and lead to models that converge to sub-optimal solutions.
\end{itemize}

\section{Additional resources}

This work includes a notebook~\cite{barschkis2023pinn} with annotations and code examples that show how to implement soft and exact BCs in PINN models. All models in this notebook are based on Keras\cite{chollet2015keras} and Tensorflow~\cite{tensorflow2015-whitepaper} as the backend. The code is open-source and readers are free to use it in their own PINN projects.

\section{Future work}

Multiple PINN instances from this study could be combined into a Mixture-of-Experts (MoE) PINN~\cite{bischof2022mixture}. I.e. a network composed of multiple child PINNs and one gating network that combines their output (e.g. through the use of a softmax function). By using such an architecture, solutions would no longer be constrained to a single BC and the Poisson equation could be solved for arbitrary BCs. Future studies should evaluate the accuracy and performance of such an MoE Poisson PINN. Additionally, a comparison with existing Poisson neural operators that can solve for arbitrary BCs should be carried out.

While the above PINN ensembles would consist of PINNs trained on different instances of the same equation (e.g. mutiple instances for Poisson non-zero BC), PINN ensembles could also consist of PINNs trained on different equations. The Laplace and Poisson zero-BC PINNs from this work, for example, could be combined into an MoE PINN that solves the Poisson non-zero BC equation. The gating function could leverage the superposition principle
\begin{align*}
    p_\text{Laplace} + p_\text{Poisson zero BC} = p_\text{Poisson non-zero BC}
\end{align*}
In general, more research should evaluate the benefits of splitting up PDEs into simpler ones, and approximating solutions with PINN ensembles.

%\begin{ack}
%\end{ack}

\nocite{*}
\small{\bibliographystyle{plainnat}\bibliography{main}}

\newpage

\appendix

\section{Training data}

Training data for experiments was generated using the method described in section~\ref{ch:data_gen}. Grid size was fixed to 128x128 cells. Except for dataset 2, all datasets used the same values ranges for variable coefficient $a$, RHS $f$, and solution at boundaries $g$. Only in dataset 2, the RHS was generated using a normalized value distribution. The purpose of dataset 2 is to evaluate if and how much normalization affects model accuracy.

\begin{table}[ht]
  \centering
  \begin{tabular}{ccccccc}
    \toprule
    Dataset \#  & \multicolumn{2}{c}{Range $a$} & \multicolumn{2}{c}{Range $f$} & \multicolumn{2}{c}{Range $p_{BC}$} \\
                & min  & max                    & min & max                     & min & max                          \\
    \midrule
    0           & -1.0 & 1.0                    & -10.0 & 10.0                  & -1.0 & 1.0                         \\
    1           & -1.0 & 1.0                    & -10.0 & 10.0                  & -1.0 & 1.0                         \\
    2           & -1.0 & 1.0                    & \textbf{-1.0} & \textbf{1.0}  & -1.0 & 1.0                         \\
    3           & -1.0 & 1.0                    & -10.0 & 10.0                  & -1.0 & 1.0                         \\
    \bottomrule\\
  \end{tabular}
  \caption{Value range of knots from datasets 0-3: Dataset 2 is the control dataset with normalized RHS.}
  \label{tab:datasets_values_ranges}
\end{table}

\newpage

\section{Poisson's equation with non-zero BC}\label{app:poisson_nonzero_exp}

\begin{figure}[ht]
\captionsetup[subfigure]{labelformat=empty,justification=centering}
\centering

\subcaptionbox{}{\includegraphics[height=.184\textwidth]{figs/eq_02/case_00/psnNet_true_eq-2_arch-3_bc-0.png}}%
\subcaptionbox{}{\includegraphics[height=.184\textwidth]{figs/eq_02/case_00/psnNet_pred_eq-2_arch-3_bc-0.png}}%
\subcaptionbox{}{\includegraphics[height=.184\textwidth]{figs/eq_02/case_00/psnNet_pred_eq-2_arch-3_bc-1.png}}%
\subcaptionbox{}{\includegraphics[height=.184\textwidth]{figs/colorbar/colorbar_1-1.png}}%
\subcaptionbox{}{\includegraphics[height=.184\textwidth]{figs/eq_02/case_00/psnNet_error_eq-2_arch-3_bc-0.png}}%
\subcaptionbox{}{\includegraphics[height=.184\textwidth]{figs/eq_02/case_00/psnNet_error_eq-2_arch-3_bc-1.png}}%
\subcaptionbox{}{\includegraphics[height=.184\textwidth]{figs/colorbar/colorbar_02-0.png}}

\subcaptionbox{}{\includegraphics[height=.184\textwidth]{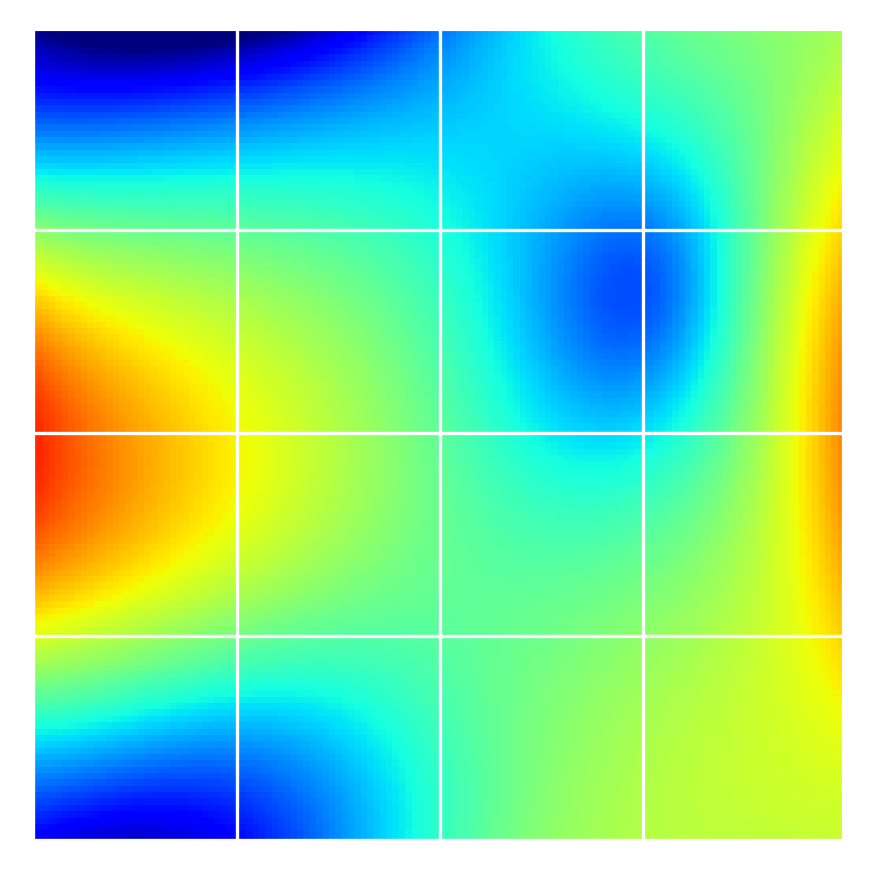}}%
\subcaptionbox{}{\includegraphics[height=.184\textwidth]{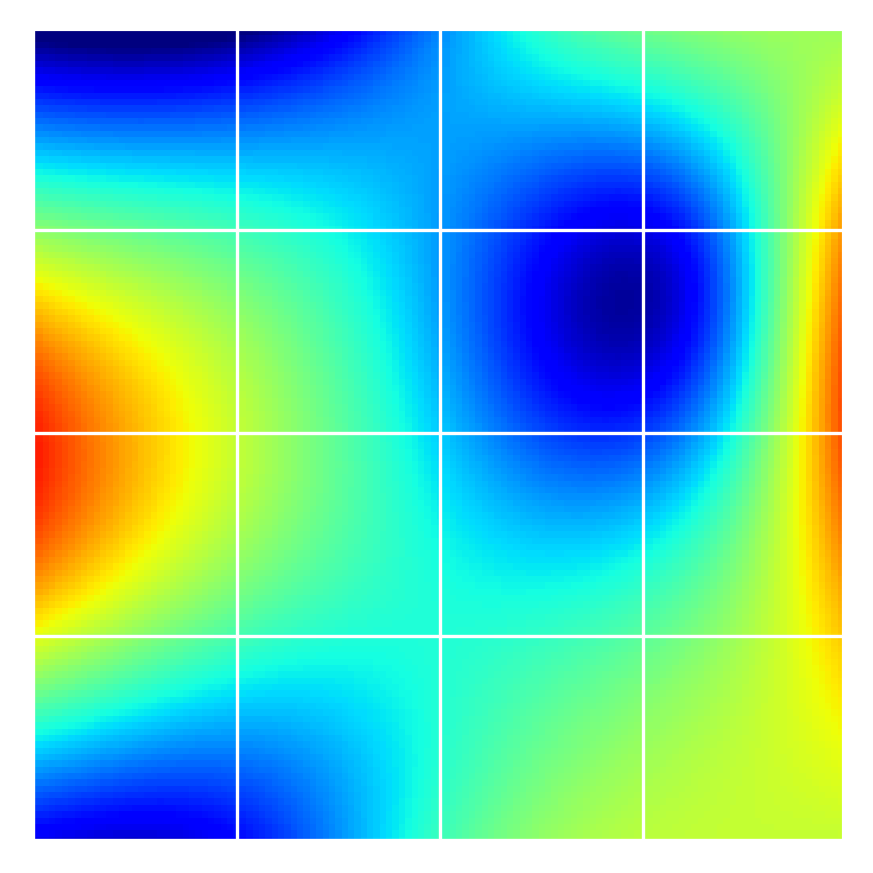}}%
\subcaptionbox{}{\includegraphics[height=.184\textwidth]{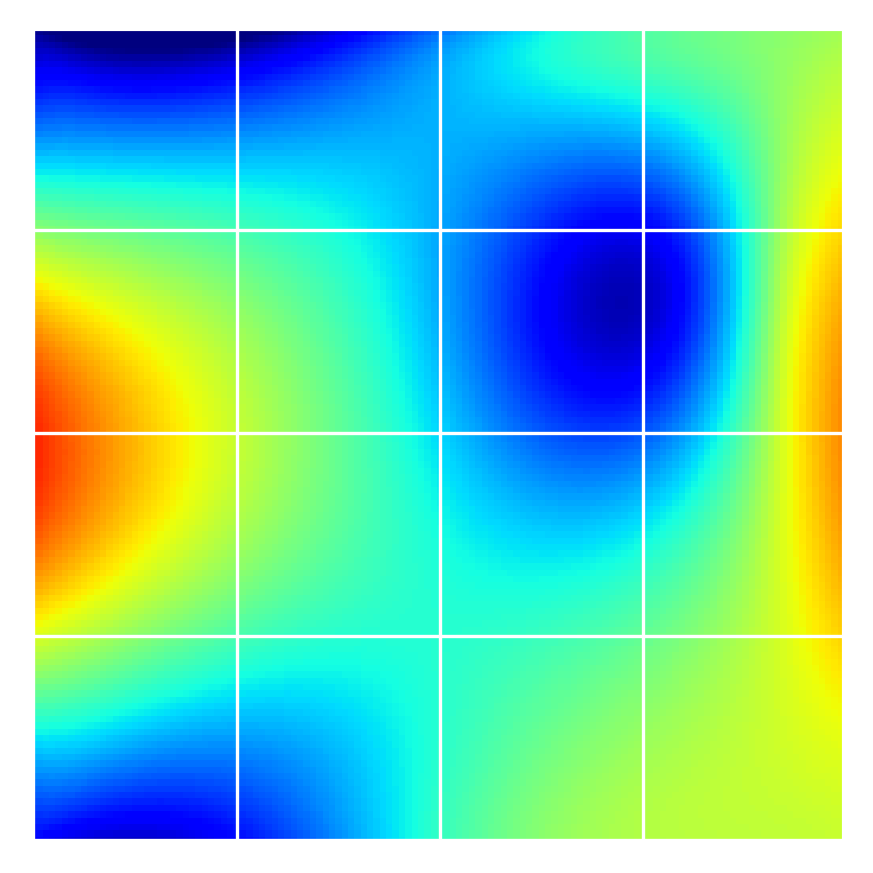}}%
\subcaptionbox{}{\includegraphics[height=.184\textwidth]{figs/colorbar/colorbar_1-1.png}}%
\subcaptionbox{}{\includegraphics[height=.184\textwidth]{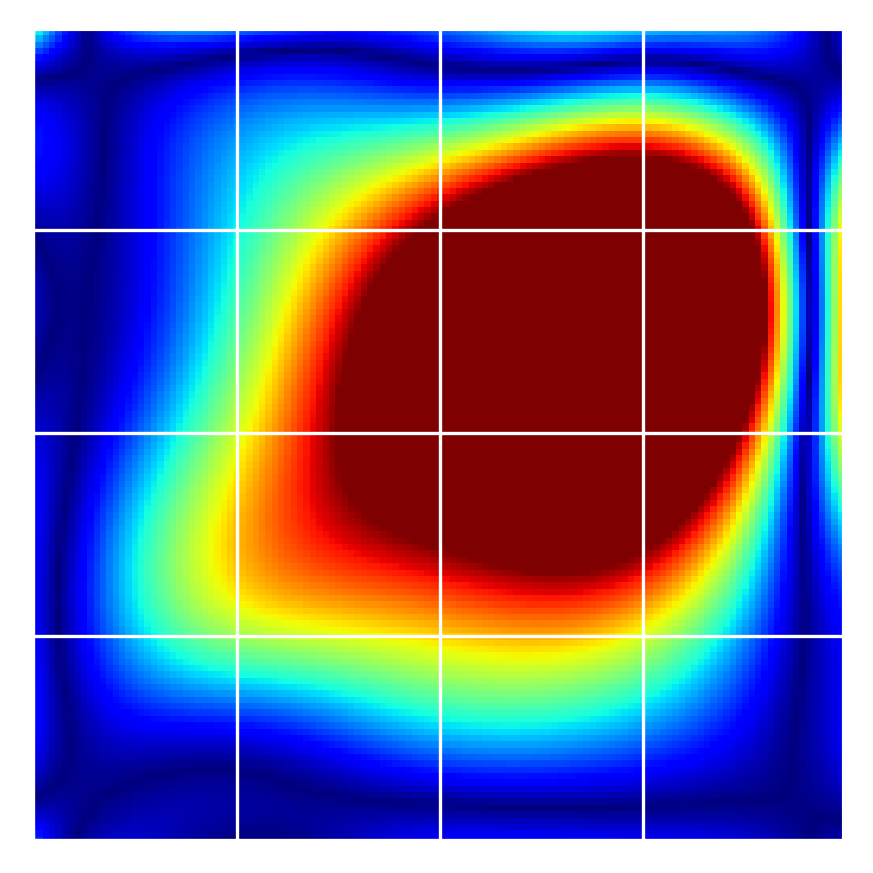}}%
\subcaptionbox{}{\includegraphics[height=.184\textwidth]{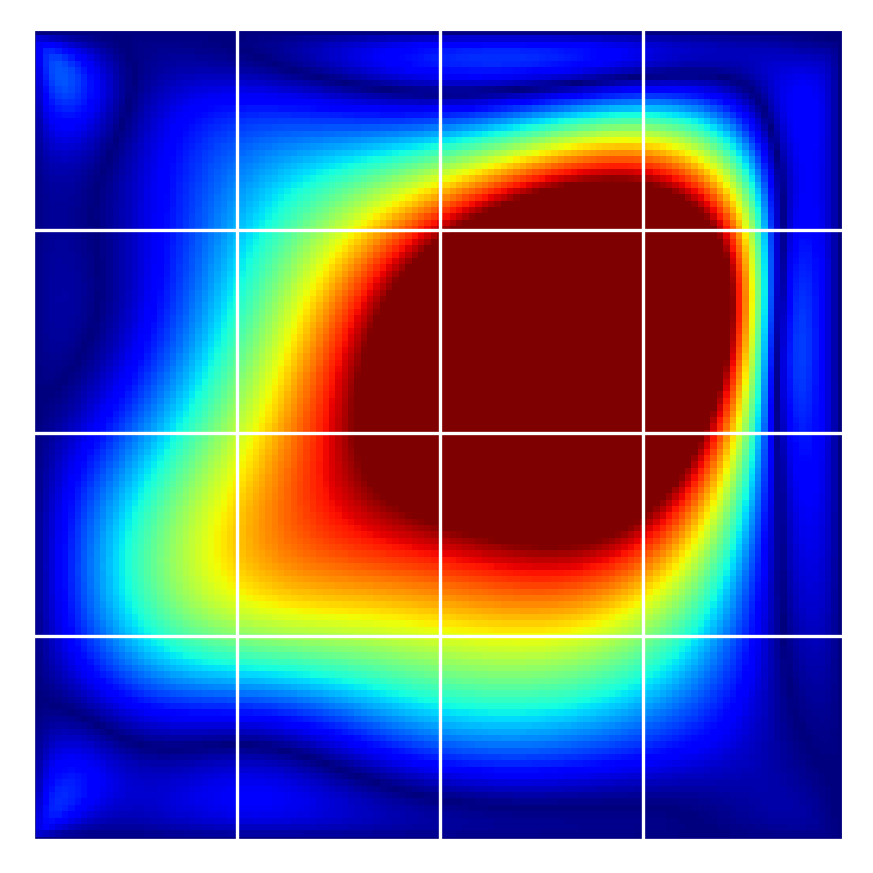}}%
\subcaptionbox{}{\includegraphics[height=.184\textwidth]{figs/colorbar/colorbar_02-0.png}}

\subcaptionbox{}{\includegraphics[height=.184\textwidth]{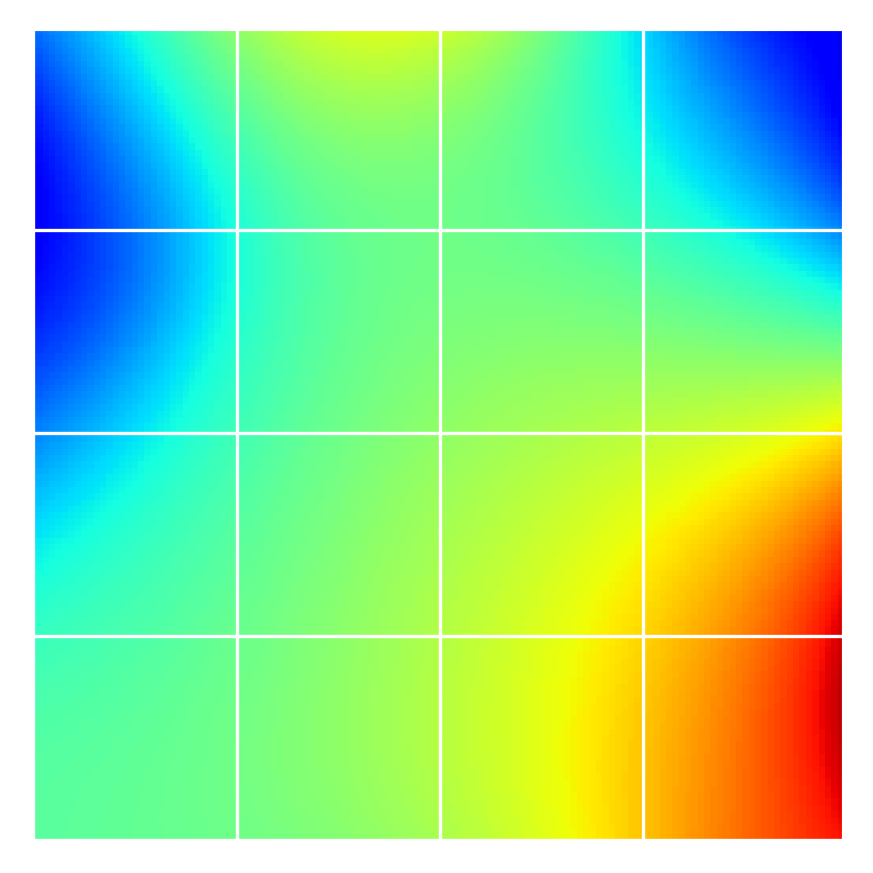}}%
\subcaptionbox{}{\includegraphics[height=.184\textwidth]{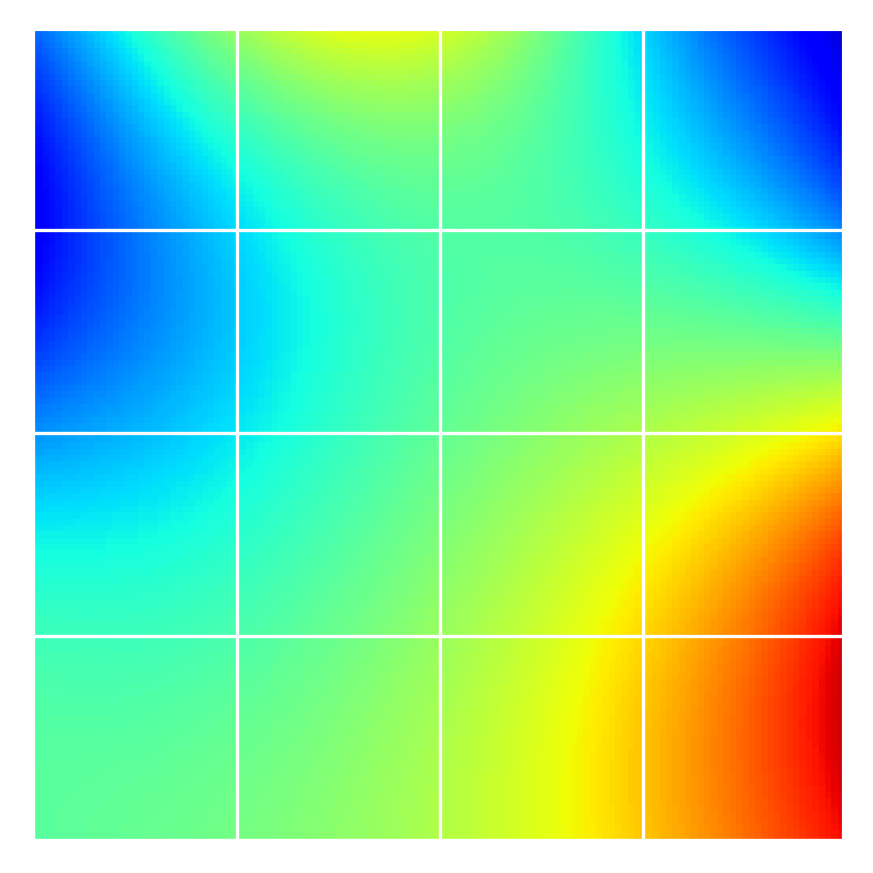}}%
\subcaptionbox{}{\includegraphics[height=.184\textwidth]{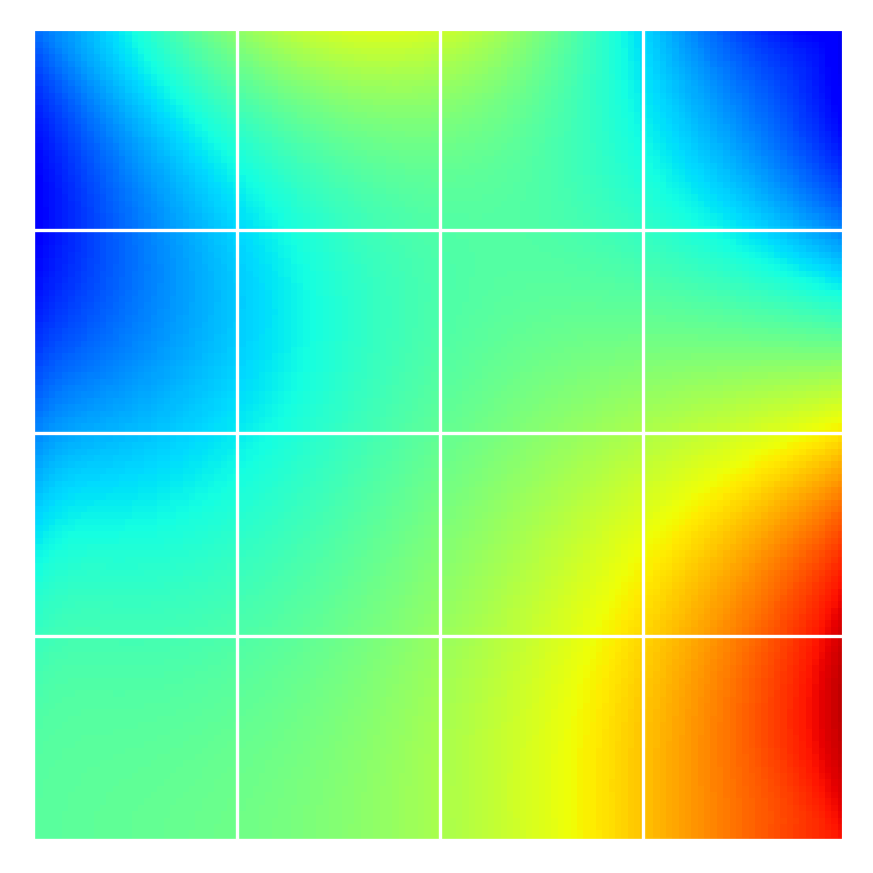}}%
\subcaptionbox{}{\includegraphics[height=.184\textwidth]{figs/colorbar/colorbar_1-1.png}}%
\subcaptionbox{}{\includegraphics[height=.184\textwidth]{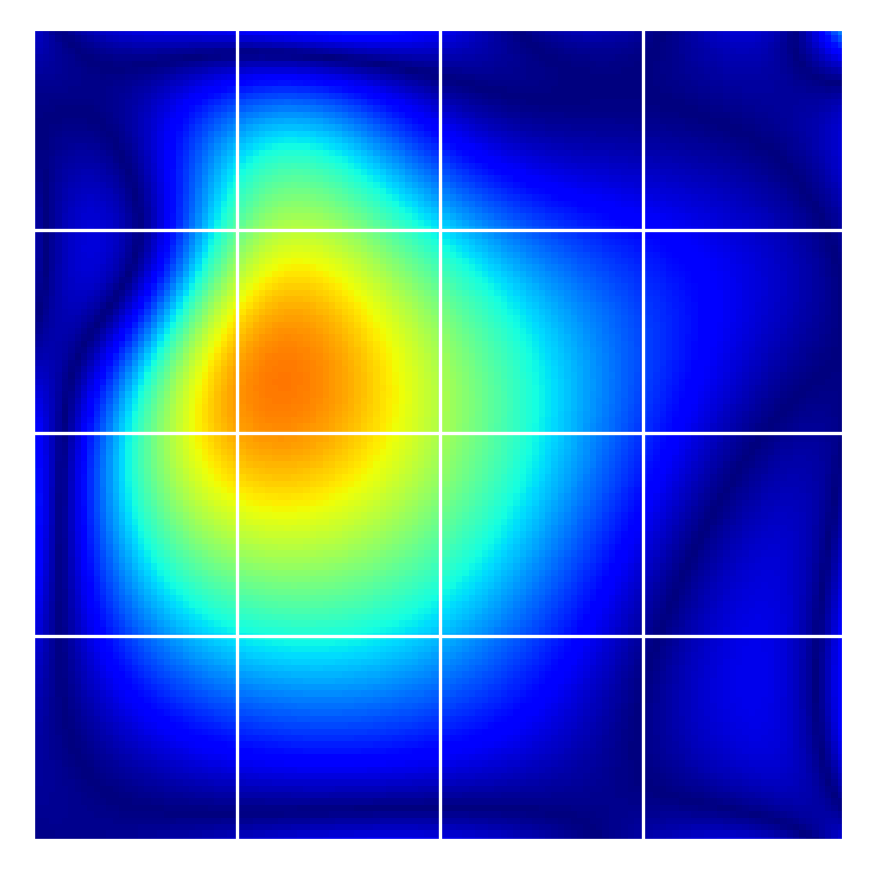}}%
\subcaptionbox{}{\includegraphics[height=.184\textwidth]{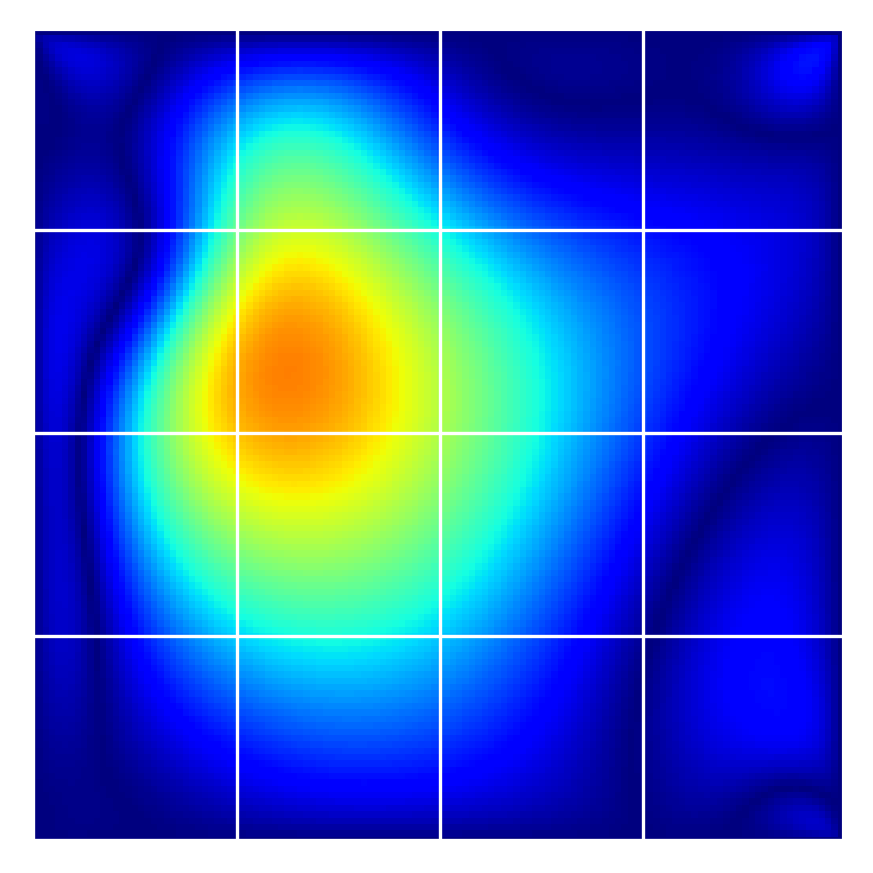}}%
\subcaptionbox{}{\includegraphics[height=.184\textwidth]{figs/colorbar/colorbar_02-0.png}}

\subcaptionbox{Ground Truth}{\includegraphics[height=.184\textwidth]{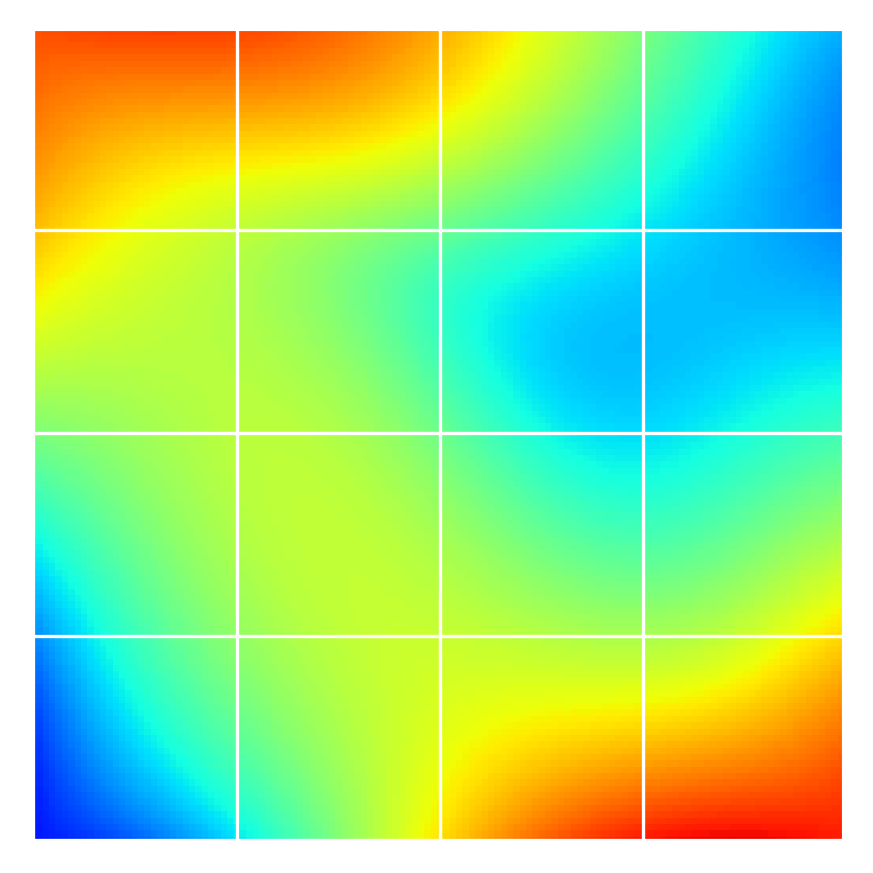}}%
\subcaptionbox{Prediction\\Soft BC}{\includegraphics[height=.184\textwidth]{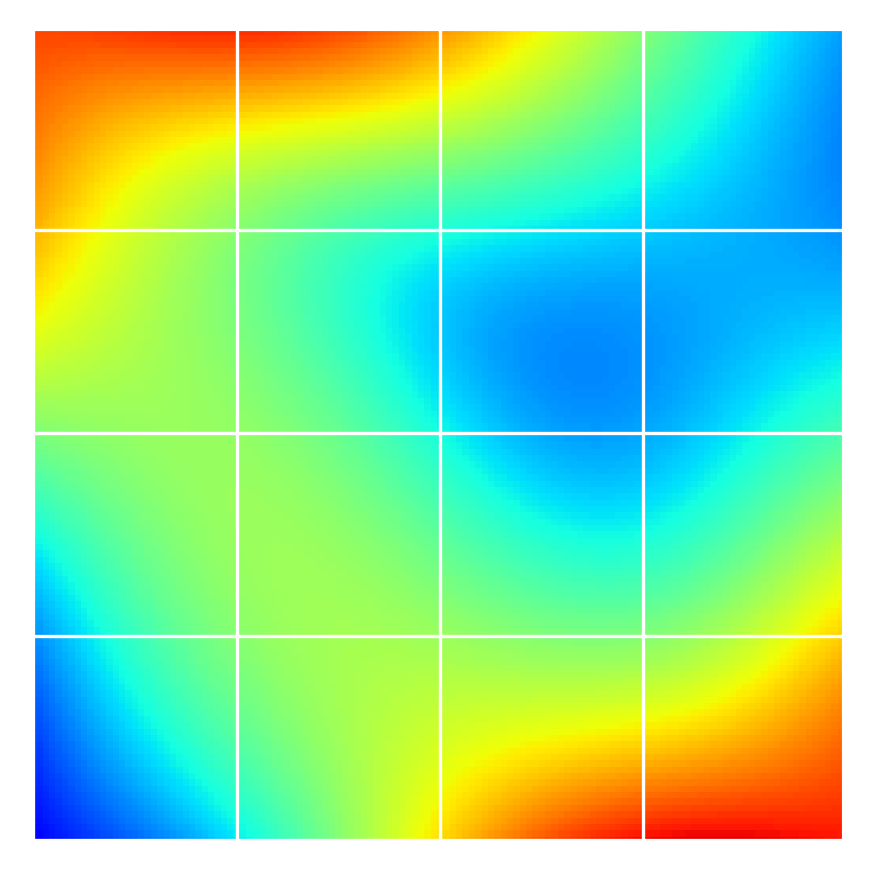}}%
\subcaptionbox{Prediction\\Exact BC}{\includegraphics[height=.184\textwidth]{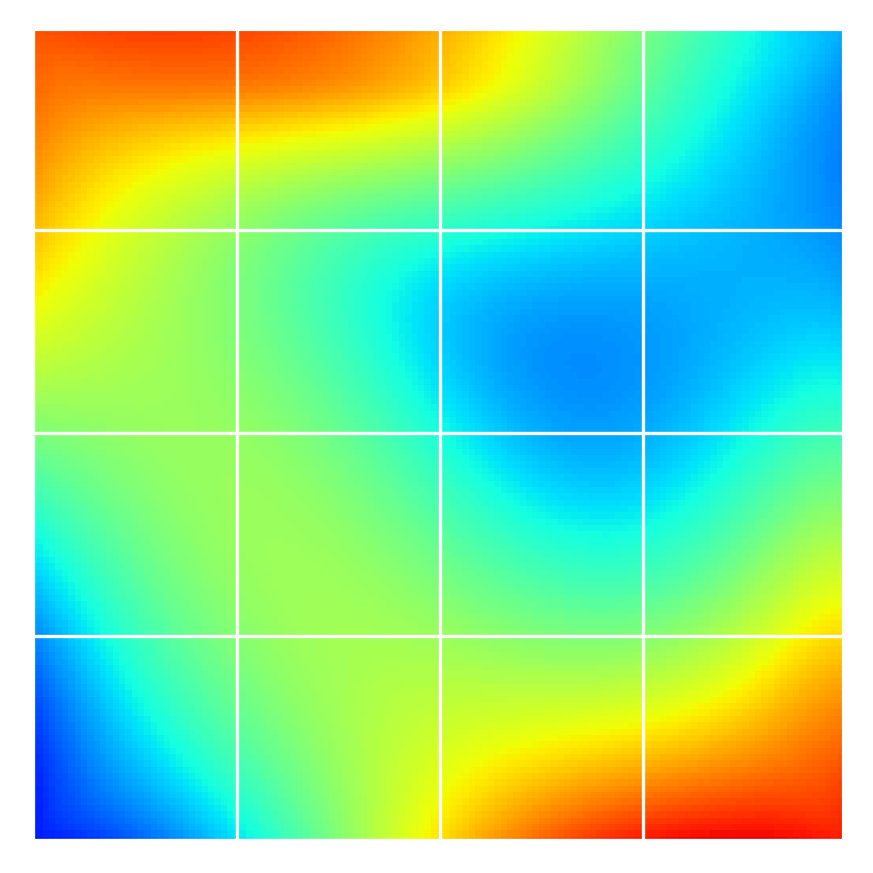}}%
\subcaptionbox{}{\includegraphics[height=.184\textwidth]{figs/colorbar/colorbar_1-1.png}}%
\subcaptionbox{Error\\Soft BC}{\includegraphics[height=.184\textwidth]{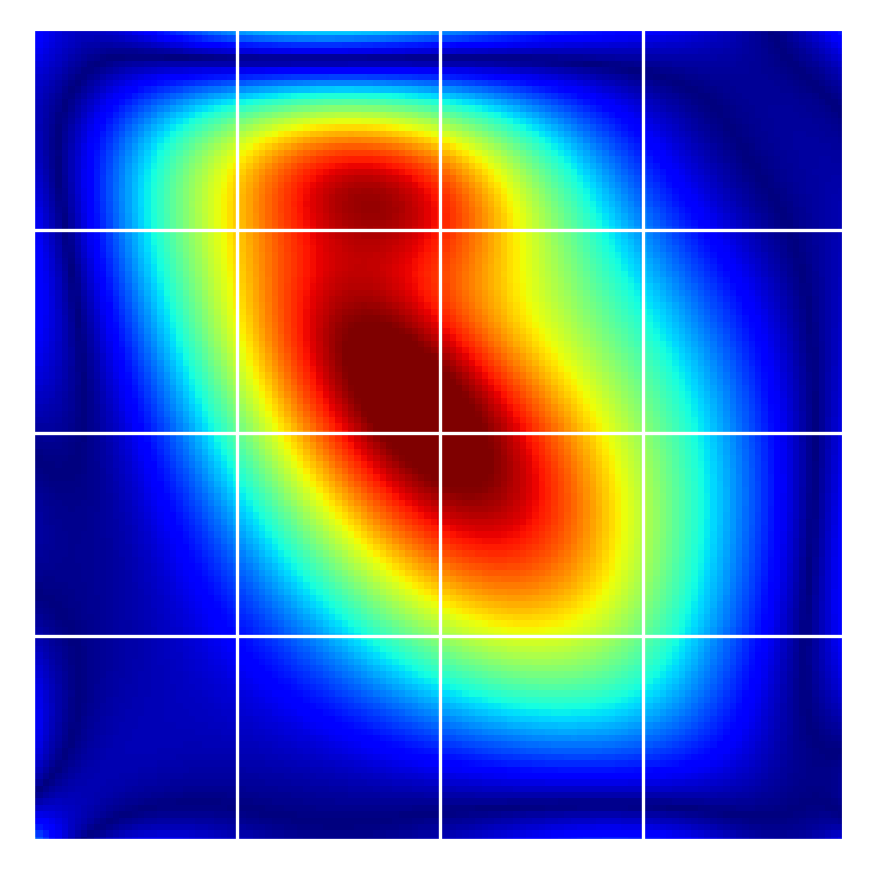}}%
\subcaptionbox{Error\\Exact BC}{\includegraphics[height=.184\textwidth]{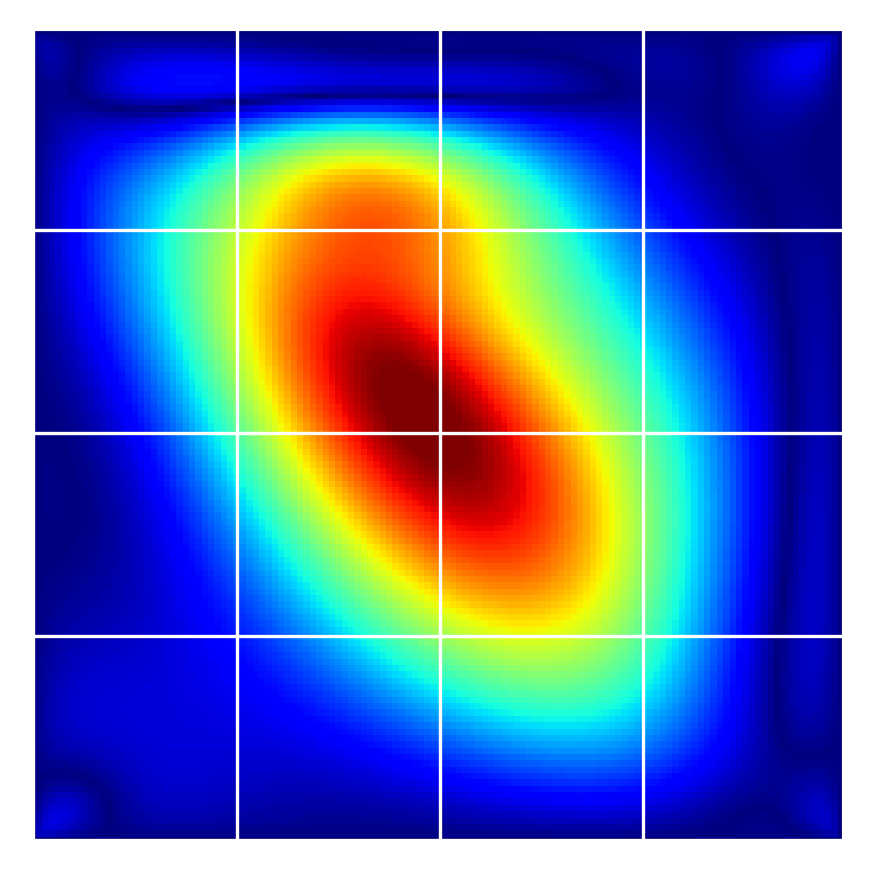}}%
\subcaptionbox{}{\includegraphics[height=.184\textwidth]{figs/colorbar/colorbar_02-0.png}}

\caption{Results from models solving Poisson's equation with non-zero BC (one dataset per row).}
\label{fig:models_eq2}
\end{figure}

\begin{table}[ht]
  \centering
  \begin{tabular}{clcccc}
    \toprule
    Dataset \#         & BC Type & MAE     & RMSE    & MAPE      & Time / epoch (sec)\\
    \midrule
    \multirow{2}{*}{0} & Soft    & 4.97e-2 & 7.01e-2 & 114.55\%  & 1.78         \\
                       & Exact   & 4.29e-2 & 6.46e-2 & 116.48\%  & 1.86         \\
    \midrule
    \multirow{2}{*}{1} & Soft    & 1.12e-1 & 1.53e-1 & 910.79\%  & 1.80         \\
                       & Exact   & 9.33e-2 & 1.32e-1 & 762.69\%  & 1.87         \\
    \midrule
    \multirow{2}{*}{2} & Soft    & 4.13e-2 & 5.92e-2 & 236.83\% & 1.79          \\
                       & Exact   & 4.19e-2 & 5.92e-2 & 245.84\% & 1.82          \\
    \midrule
    \multirow{2}{*}{3} & Soft    & 6.64e-2 & 9.14e-2 & 438.13\% & 1.79          \\
                       & Exact   & 5.86e-2 & 8.29e-2 & 397.74\% & 1.88          \\
    \bottomrule\\
  \end{tabular}
  \caption{Error metrics for models solving Poisson's equation with non-zero BC (plots in Figure~\ref{fig:models_eq2})}
  \label{tab:models_eq2}
\end{table}

\newpage

\section{Laplace's equation}\label{app:laplace_exp}

\begin{figure}[ht]
\captionsetup[subfigure]{labelformat=empty,justification=centering}
\centering

\subcaptionbox{}{\includegraphics[height=.184\textwidth]{figs/eq_01/case_00/psnNet_true_eq-1_arch-3_bc-0.png}}%
\subcaptionbox{}{\includegraphics[height=.184\textwidth]{figs/eq_01/case_00/psnNet_pred_eq-1_arch-3_bc-0.png}}%
\subcaptionbox{}{\includegraphics[height=.184\textwidth]{figs/eq_01/case_00/psnNet_pred_eq-1_arch-3_bc-1.png}}%
\subcaptionbox{}{\includegraphics[height=.184\textwidth]{figs/colorbar/colorbar_1-1.png}}%
\subcaptionbox{}{\includegraphics[height=.184\textwidth]{figs/eq_01/case_00/psnNet_error_eq-1_arch-3_bc-0.png}}%
\subcaptionbox{}{\includegraphics[height=.184\textwidth]{figs/eq_01/case_00/psnNet_error_eq-1_arch-3_bc-1.png}}%
\subcaptionbox{}{\includegraphics[height=.184\textwidth]{figs/colorbar/colorbar_02-0.png}}

\subcaptionbox{}{\includegraphics[height=.184\textwidth]{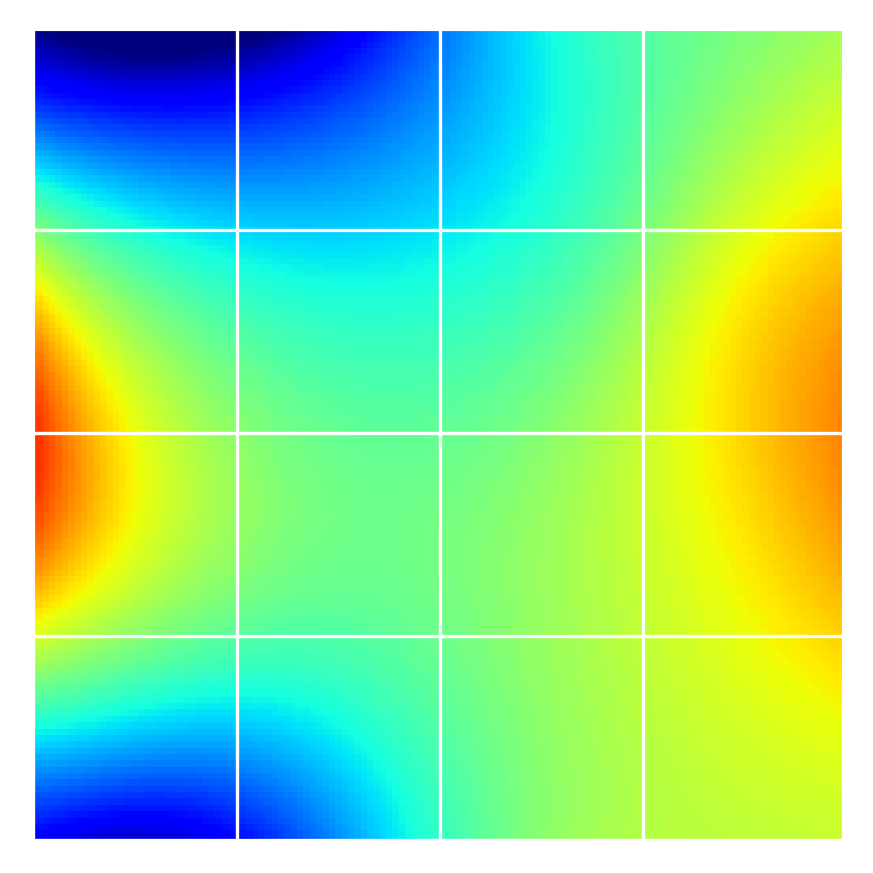}}%
\subcaptionbox{}{\includegraphics[height=.184\textwidth]{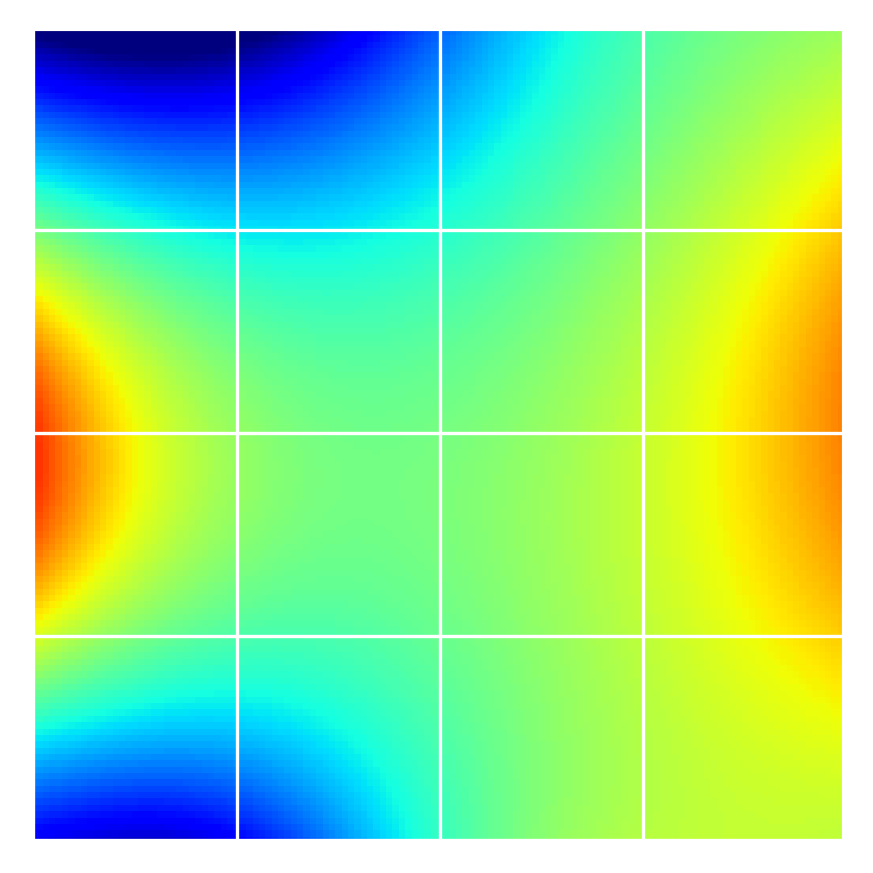}}%
\subcaptionbox{}{\includegraphics[height=.184\textwidth]{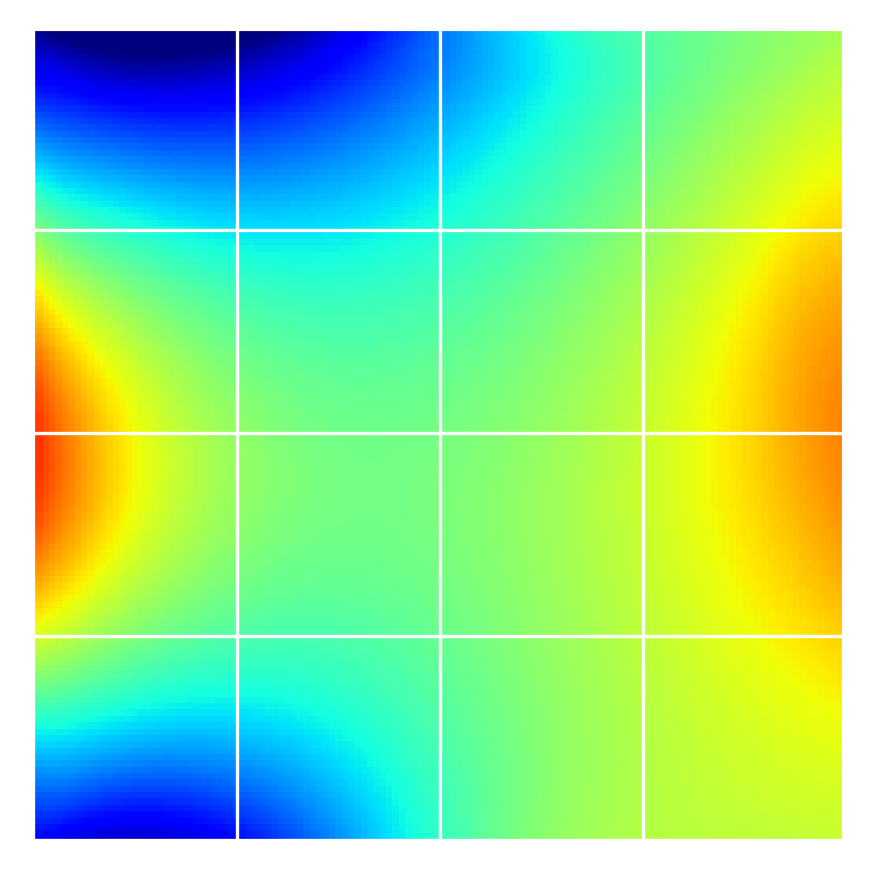}}%
\subcaptionbox{}{\includegraphics[height=.184\textwidth]{figs/colorbar/colorbar_1-1.png}}%
\subcaptionbox{}{\includegraphics[height=.184\textwidth]{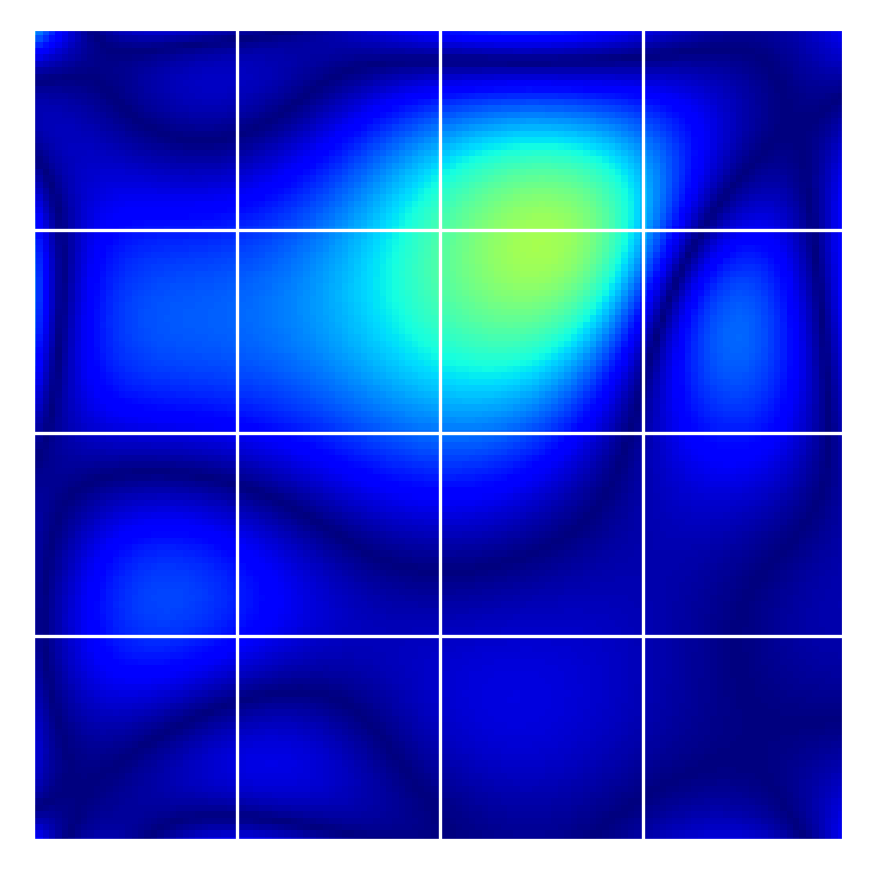}}%
\subcaptionbox{}{\includegraphics[height=.184\textwidth]{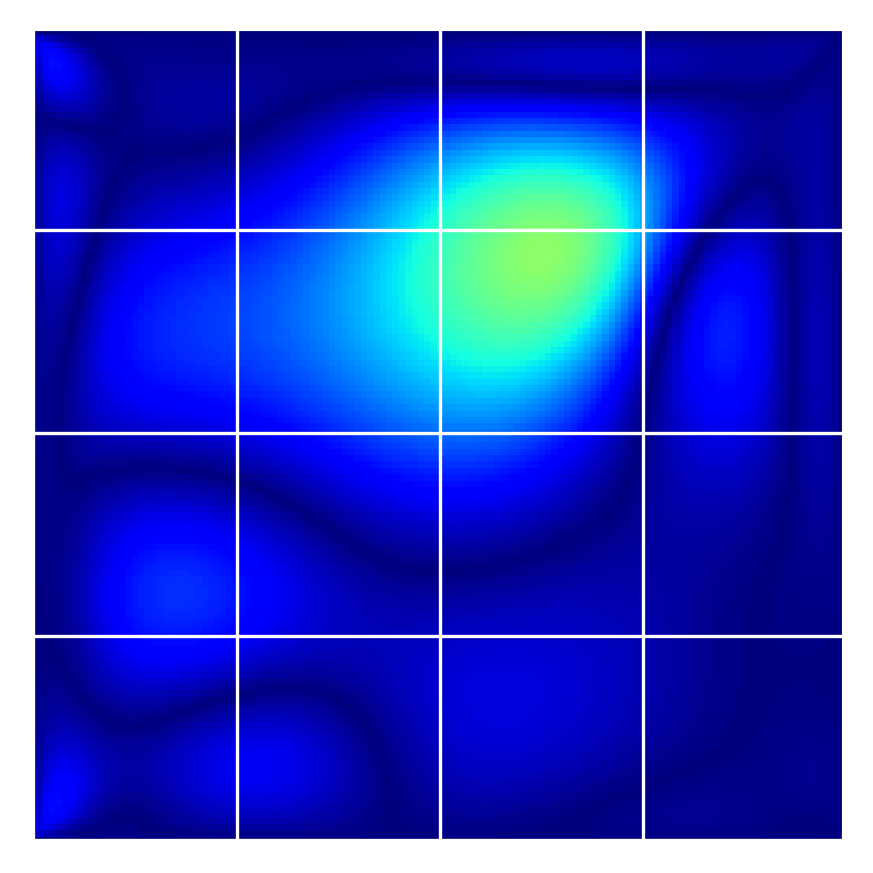}}%
\subcaptionbox{}{\includegraphics[height=.184\textwidth]{figs/colorbar/colorbar_02-0.png}}

\subcaptionbox{}{\includegraphics[height=.184\textwidth]{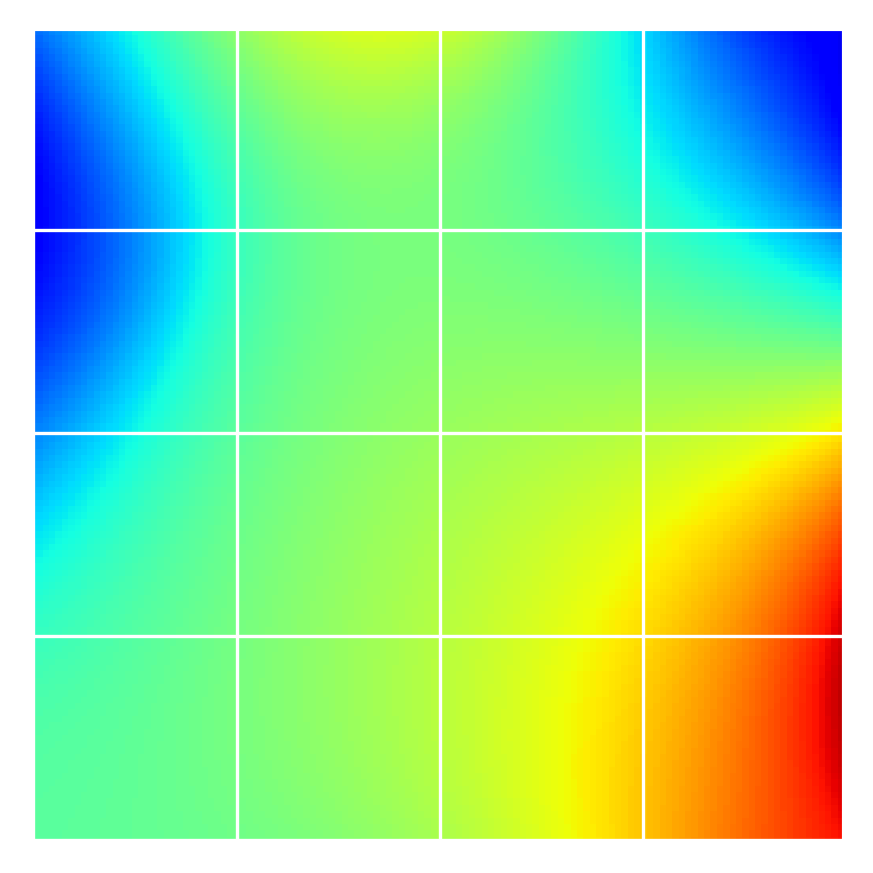}}%
\subcaptionbox{}{\includegraphics[height=.184\textwidth]{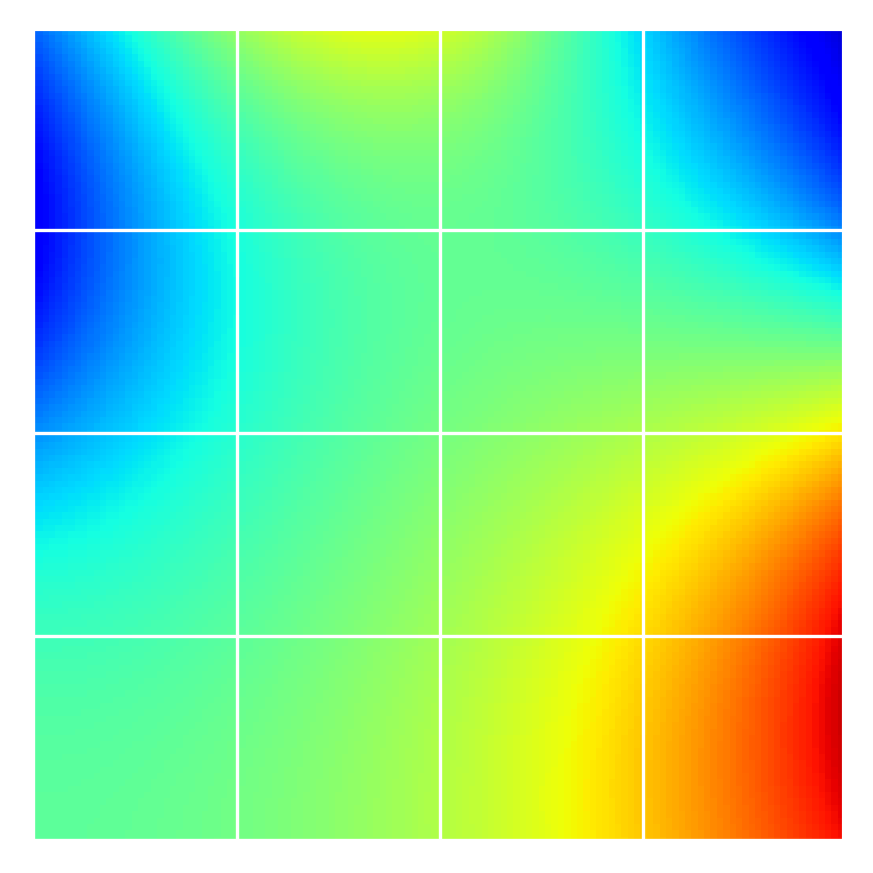}}%
\subcaptionbox{}{\includegraphics[height=.184\textwidth]{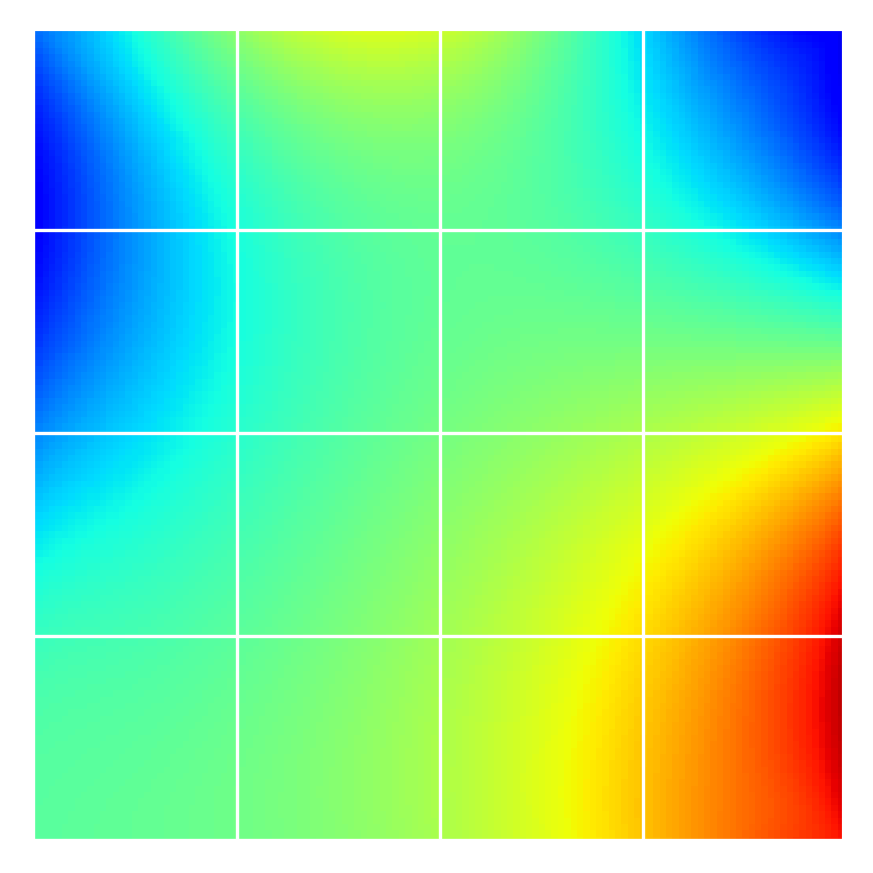}}%
\subcaptionbox{}{\includegraphics[height=.184\textwidth]{figs/colorbar/colorbar_1-1.png}}%
\subcaptionbox{}{\includegraphics[height=.184\textwidth]{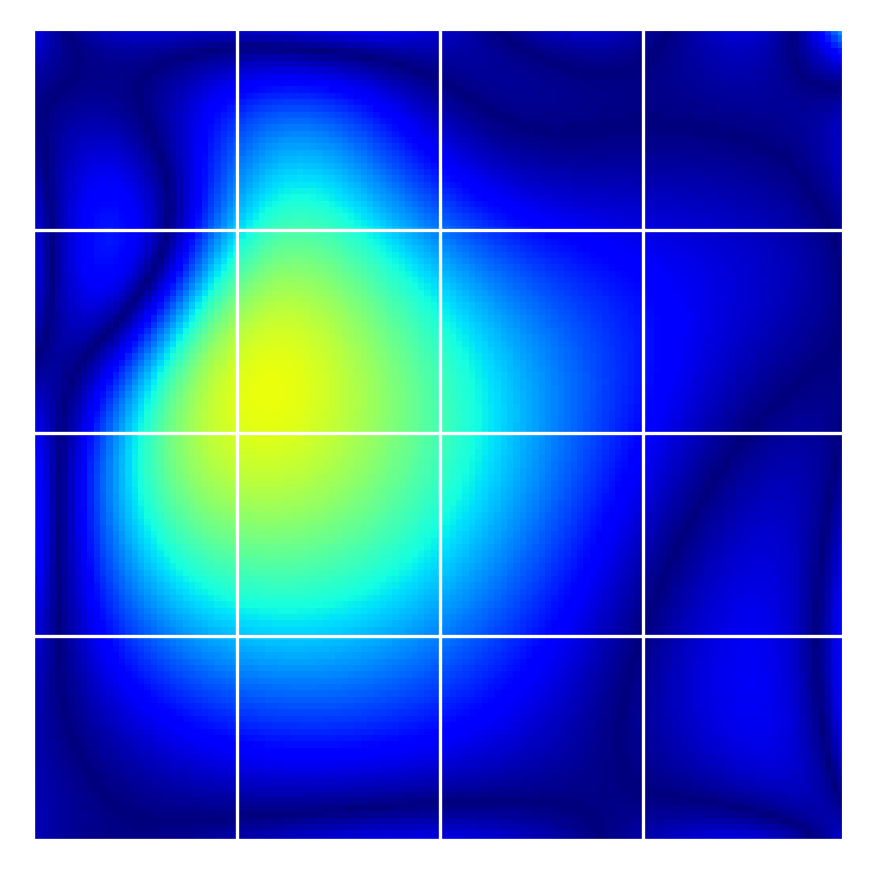}}%
\subcaptionbox{}{\includegraphics[height=.184\textwidth]{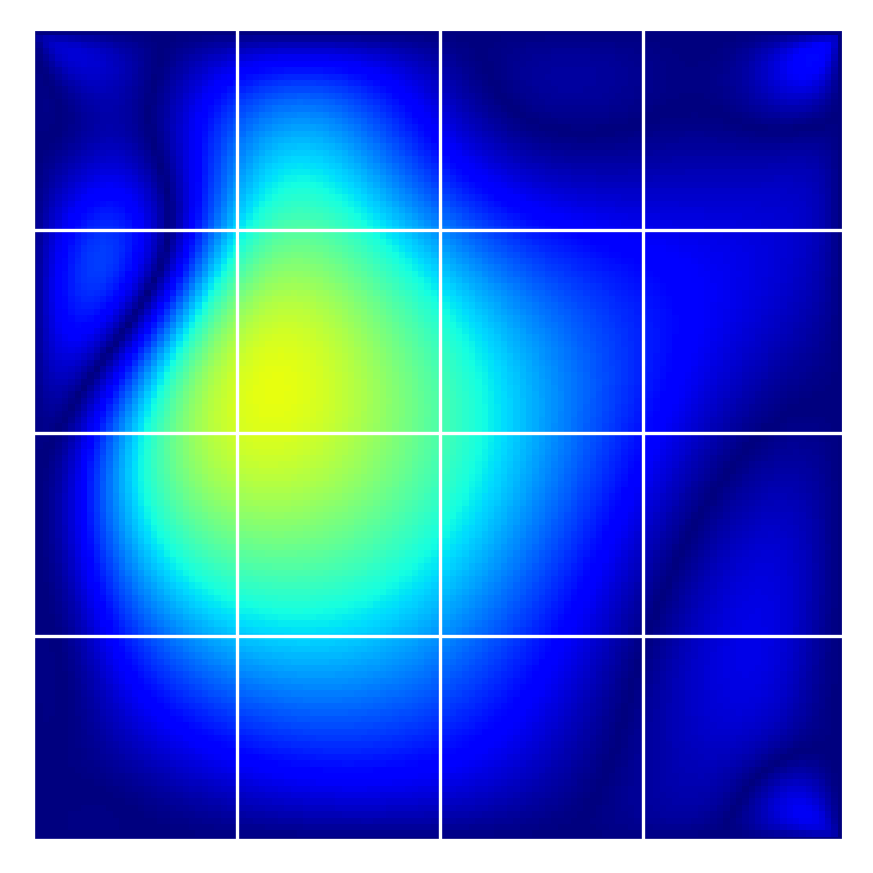}}%
\subcaptionbox{}{\includegraphics[height=.184\textwidth]{figs/colorbar/colorbar_02-0.png}}

\subcaptionbox{Ground Truth}{\includegraphics[height=.184\textwidth]{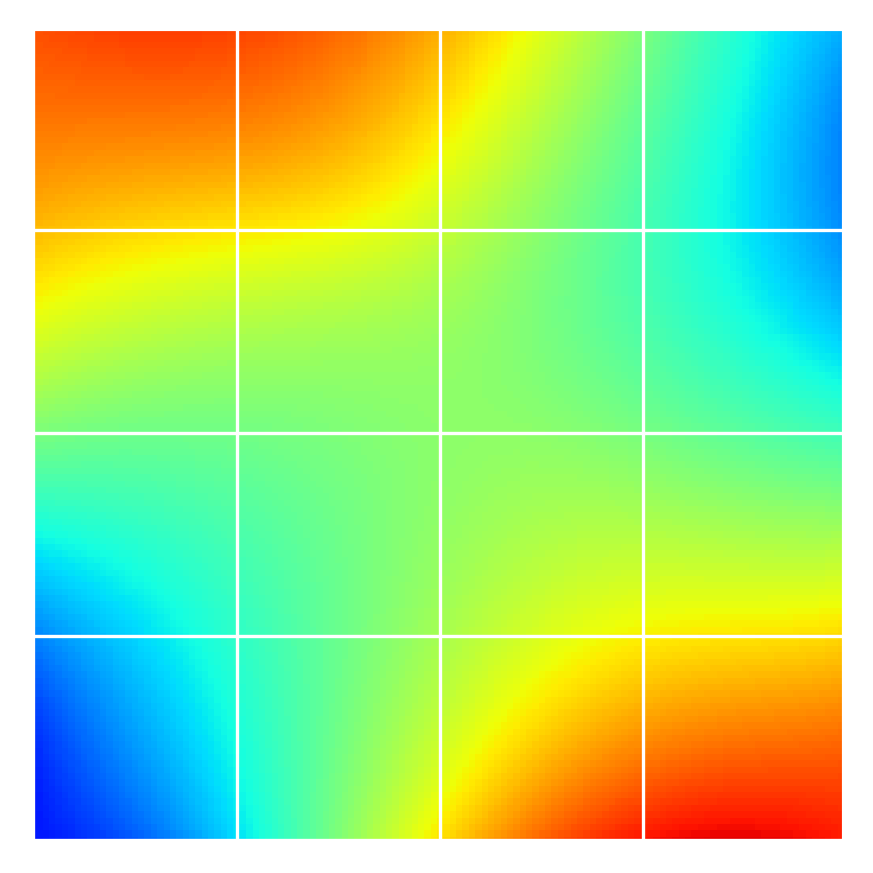}}%
\subcaptionbox{Prediction Soft BC}{\includegraphics[height=.184\textwidth]{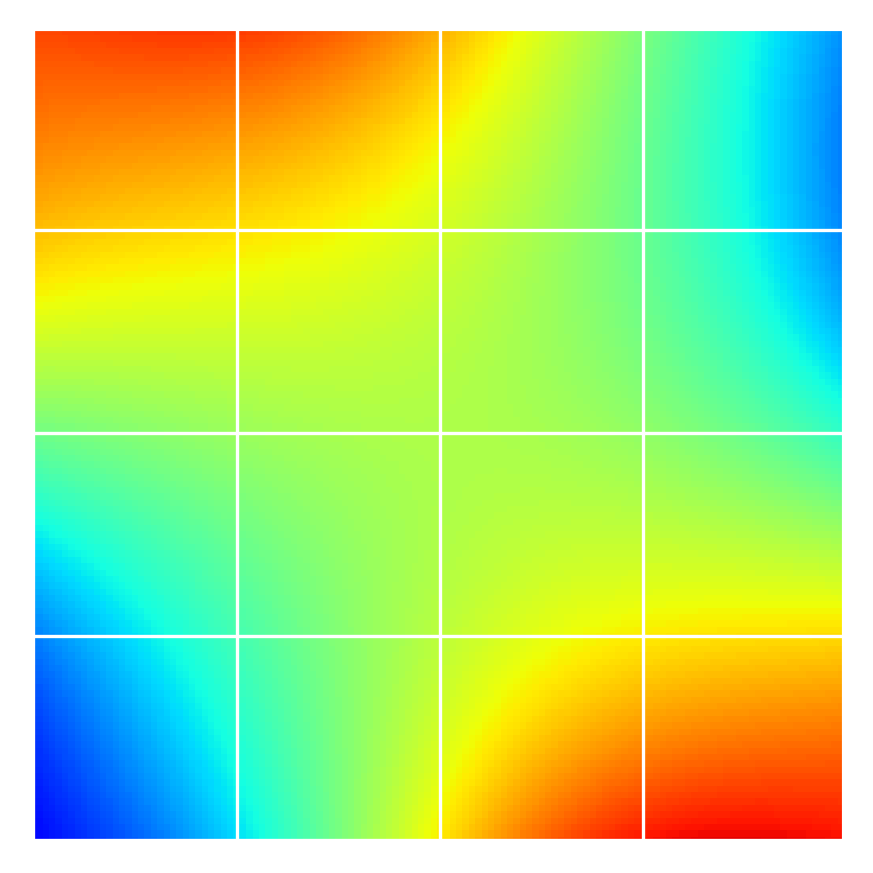}}%
\subcaptionbox{Prediction Exact BC}{\includegraphics[height=.184\textwidth]{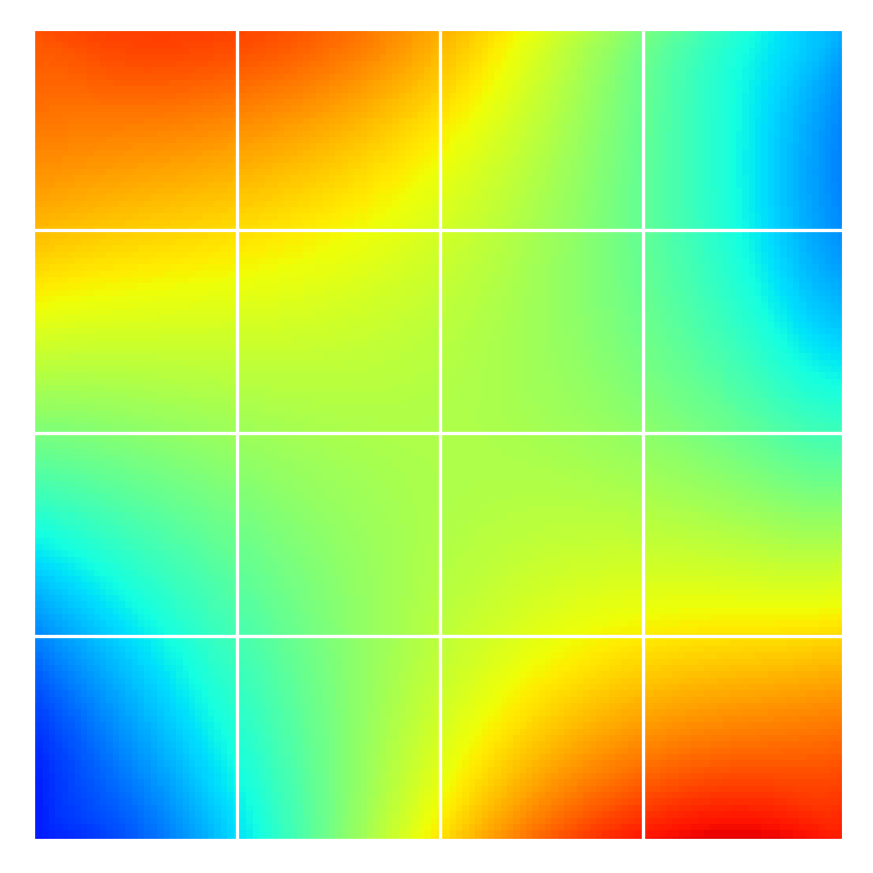}}%
\subcaptionbox{}{\includegraphics[height=.184\textwidth]{figs/colorbar/colorbar_1-1.png}}%
\subcaptionbox{Error Soft BC}{\includegraphics[height=.184\textwidth]{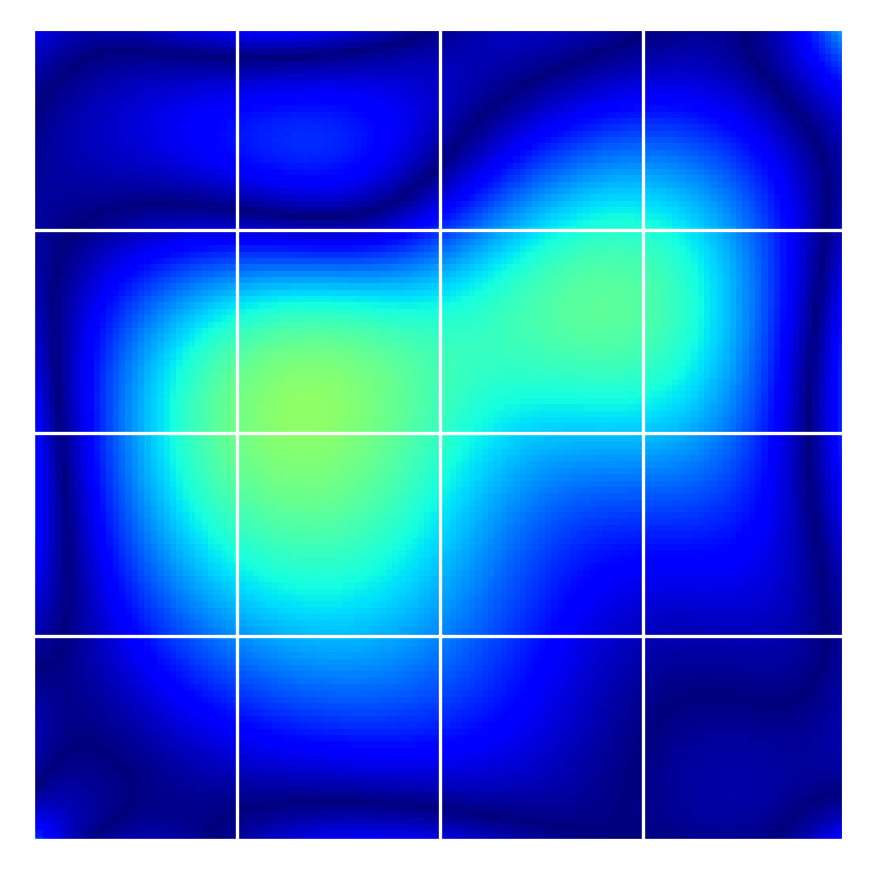}}%
\subcaptionbox{Error Exact BC}{\includegraphics[height=.184\textwidth]{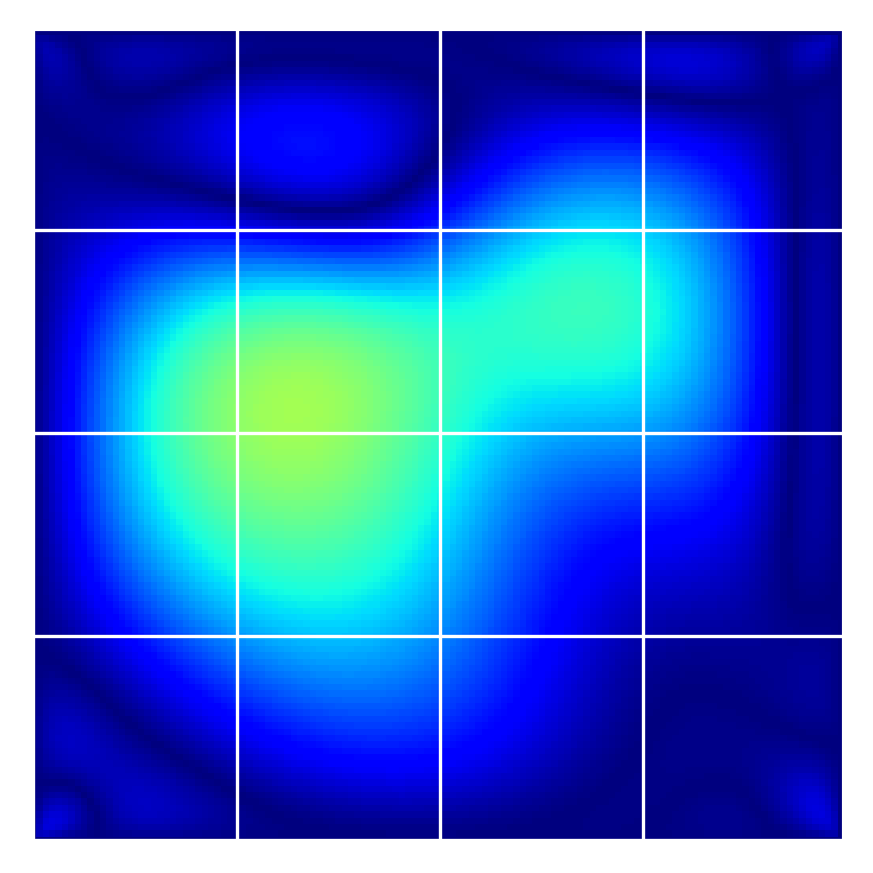}}%
\subcaptionbox{}{\includegraphics[height=.184\textwidth]{figs/colorbar/colorbar_02-0.png}}

\caption{Results from models solving Laplace's equation with non-zero BC (one dataset per row).}
\label{fig:models_eq1}
\end{figure}

\begin{table}[ht]
  \centering
  \begin{tabular}{clcccc}
    \toprule
    Dataset \#         & BC Type & MAE     & RMSE    & MAPE      & Avg. time / epoch (sec)\\
    \midrule
    \multirow{2}{*}{0} & Soft    & 2.53e-2 & 3.19e-2 & 150.71\%  & 1.79         \\
                       & Exact   & 2.19e-2 & 2.99e-2 & 140.02\%  & 1.84         \\
    \midrule
    \multirow{2}{*}{1} & Soft    & 2.27e-2 & 3.33e-2 & 91.42\%  & 1.81         \\
                       & Exact   & 1.88e-2 & 2.97e-2 & 84.89\%  & 1.84         \\
    \midrule
    \multirow{2}{*}{2} & Soft    & 3.31e-2 & 4.73e-2 & 236.63\% & 1.85          \\
                       & Exact   & 3.42e-2 & 4.82e-2 & 246.42\% & 1.85          \\
    \midrule
    \multirow{2}{*}{3} & Soft    & 3.52e-2 & 4.63e-2 & 271.54\% & 1.78          \\
                       & Exact   & 3.30e-2 & 4.55e-2 & 276.54\% & 1.84          \\
    \bottomrule\\
  \end{tabular}
  \caption{Error metrics for models solving Laplace's equation with non-zero BC (plots in Figure~\ref{fig:models_eq1})}
  \label{tab:models_eq1}
\end{table}

\newpage

\section{Poisson's equation with zero BC}\label{app:poisson_zero_exp}

\begin{figure}[ht]
\captionsetup[subfigure]{labelformat=empty,justification=centering}
\centering

\subcaptionbox{}{\includegraphics[height=.184\textwidth]{figs/eq_00/case_00/psnNet_true_eq-0_arch-3_bc-0.png}}%
\subcaptionbox{}{\includegraphics[height=.184\textwidth]{figs/eq_00/case_00/psnNet_pred_eq-0_arch-3_bc-0.png}}%
\subcaptionbox{}{\includegraphics[height=.184\textwidth]{figs/eq_00/case_00/psnNet_pred_eq-0_arch-3_bc-1.png}}%
\subcaptionbox{}{\includegraphics[height=.184\textwidth]{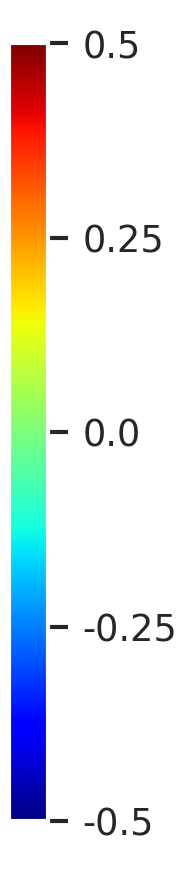}}%
\subcaptionbox{}{\includegraphics[height=.184\textwidth]{figs/eq_00/case_00/psnNet_error_eq-0_arch-3_bc-0.png}}%
\subcaptionbox{}{\includegraphics[height=.184\textwidth]{figs/eq_00/case_00/psnNet_error_eq-0_arch-3_bc-1.png}}%
\subcaptionbox{}{\includegraphics[height=.184\textwidth]{figs/colorbar/colorbar_02-0.png}}

\subcaptionbox{}{\includegraphics[height=.184\textwidth]{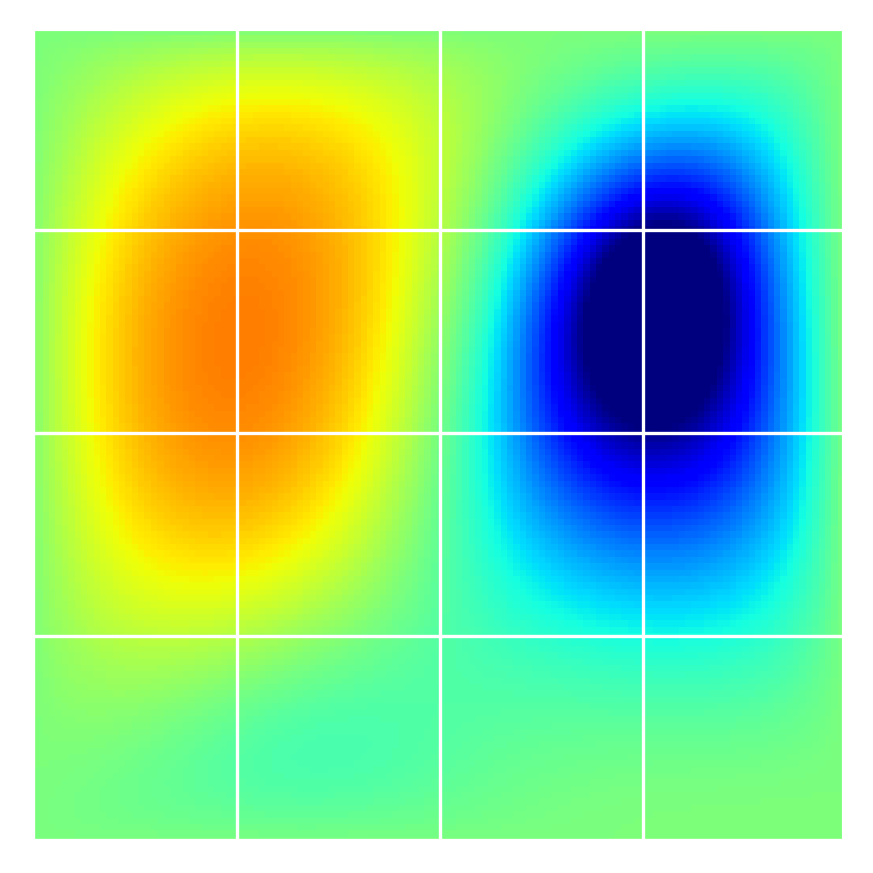}}%
\subcaptionbox{}{\includegraphics[height=.184\textwidth]{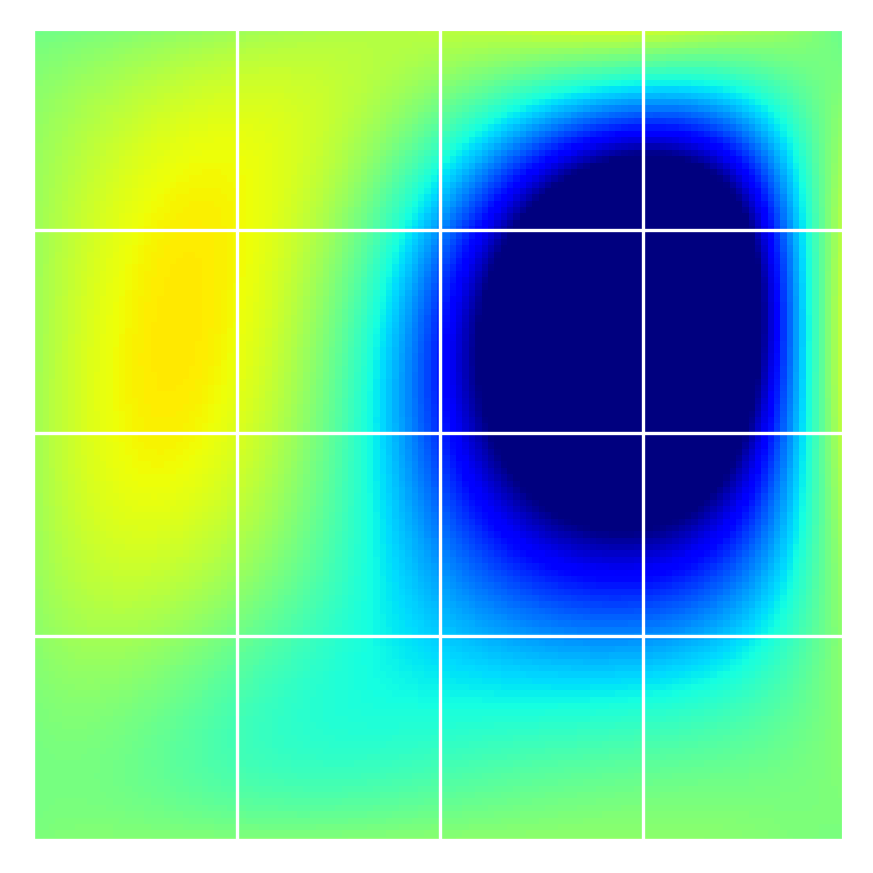}}%
\subcaptionbox{}{\includegraphics[height=.184\textwidth]{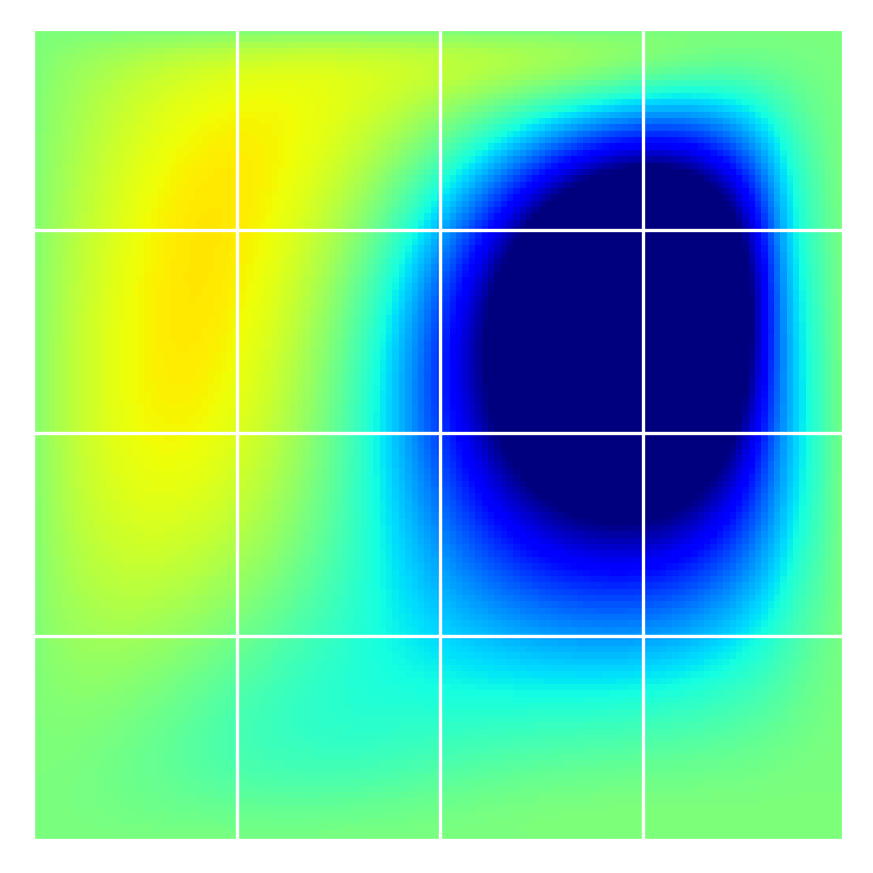}}%
\subcaptionbox{}{\includegraphics[height=.184\textwidth]{figs/colorbar/colorbar_05-05.png}}%
\subcaptionbox{}{\includegraphics[height=.184\textwidth]{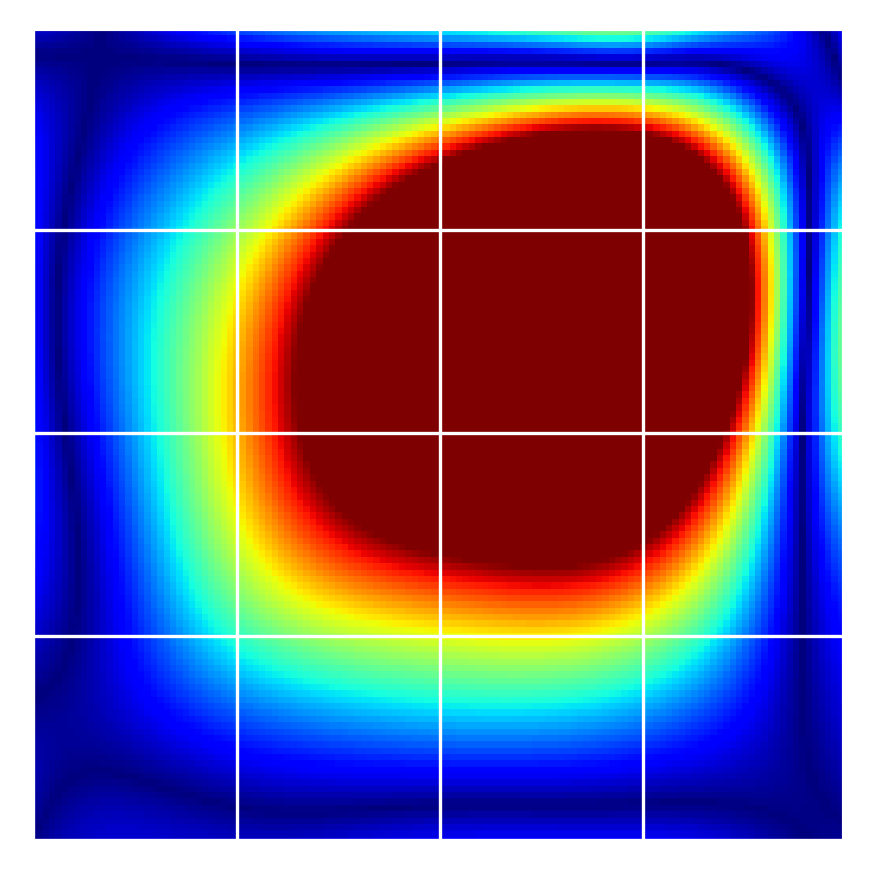}}%
\subcaptionbox{}{\includegraphics[height=.184\textwidth]{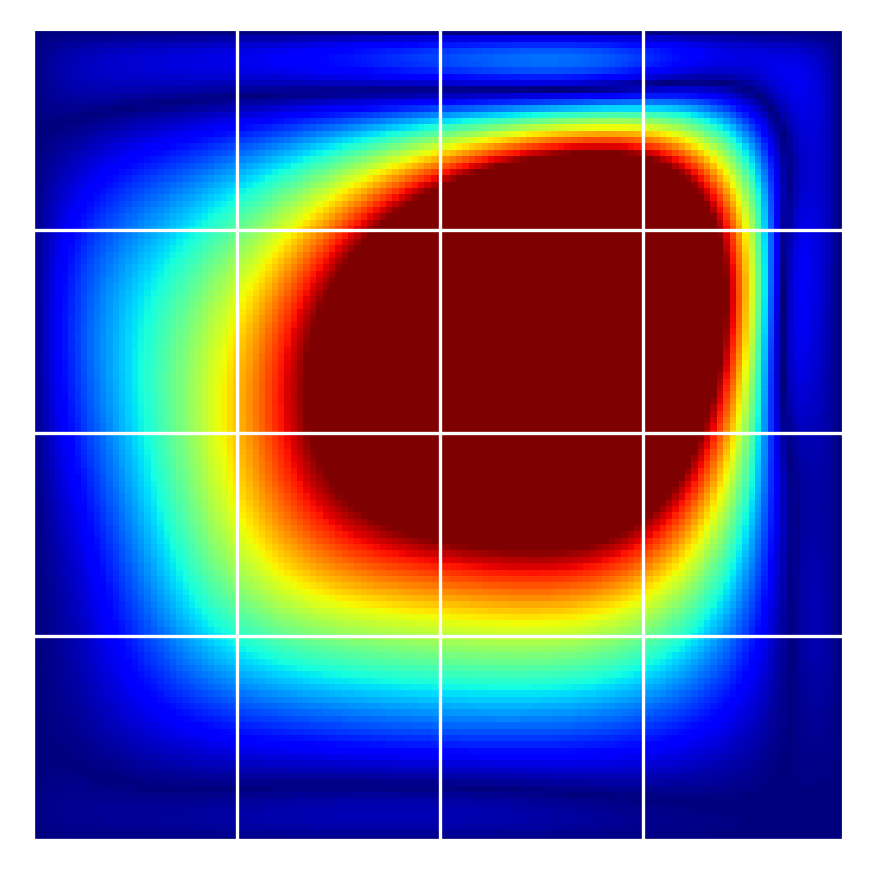}}%
\subcaptionbox{}{\includegraphics[height=.184\textwidth]{figs/colorbar/colorbar_02-0.png}}

\subcaptionbox{}{\includegraphics[height=.184\textwidth]{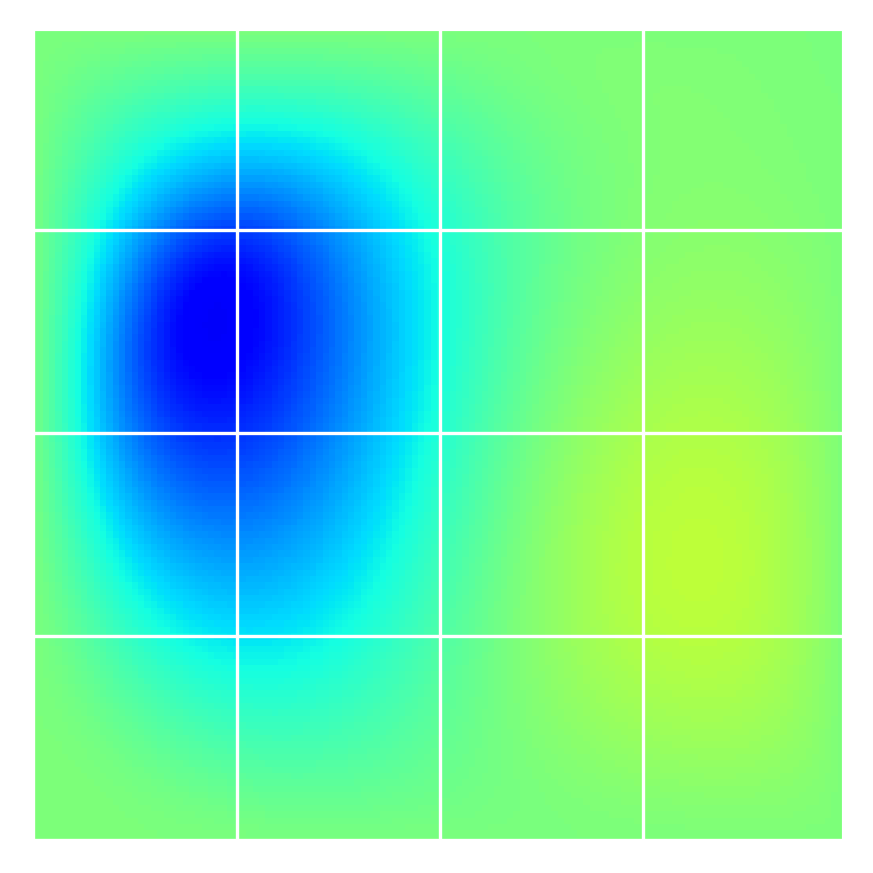}}%
\subcaptionbox{}{\includegraphics[height=.184\textwidth]{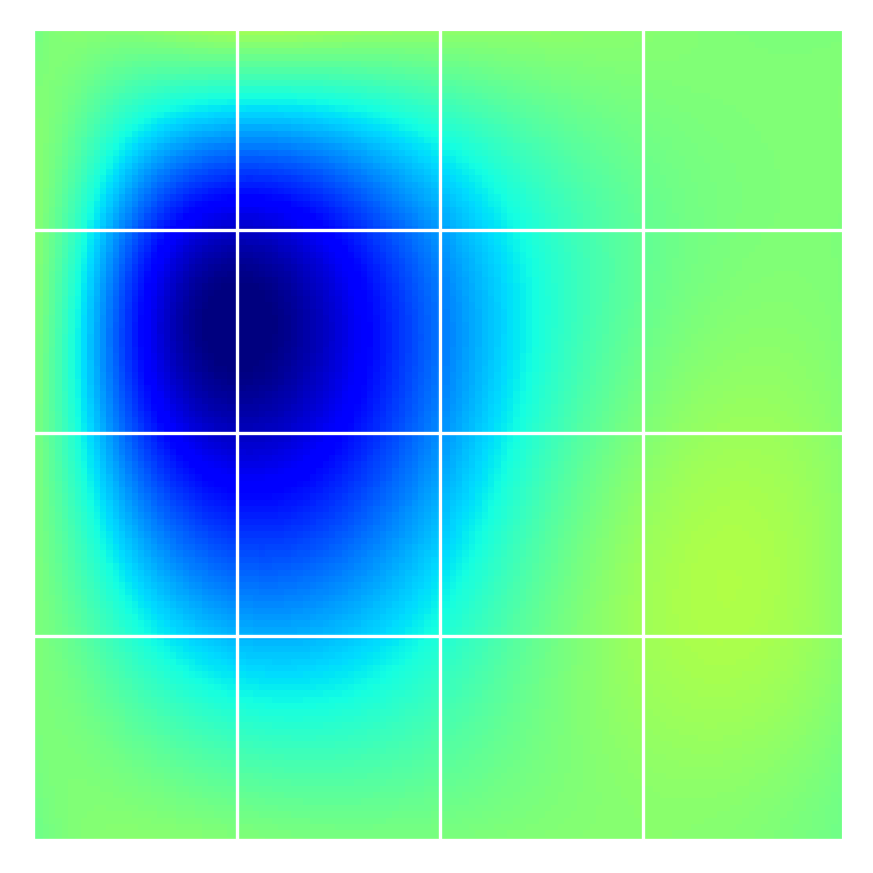}}%
\subcaptionbox{}{\includegraphics[height=.184\textwidth]{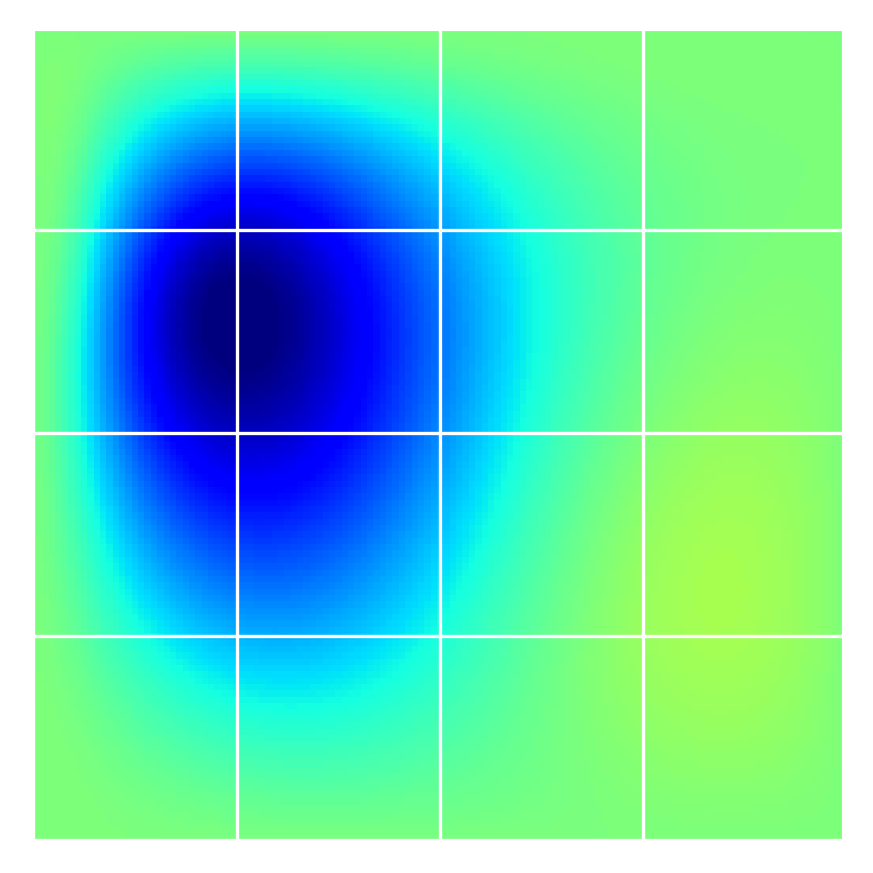}}%
\subcaptionbox{}{\includegraphics[height=.184\textwidth]{figs/colorbar/colorbar_01-01.png}}%
\subcaptionbox{}{\includegraphics[height=.184\textwidth]{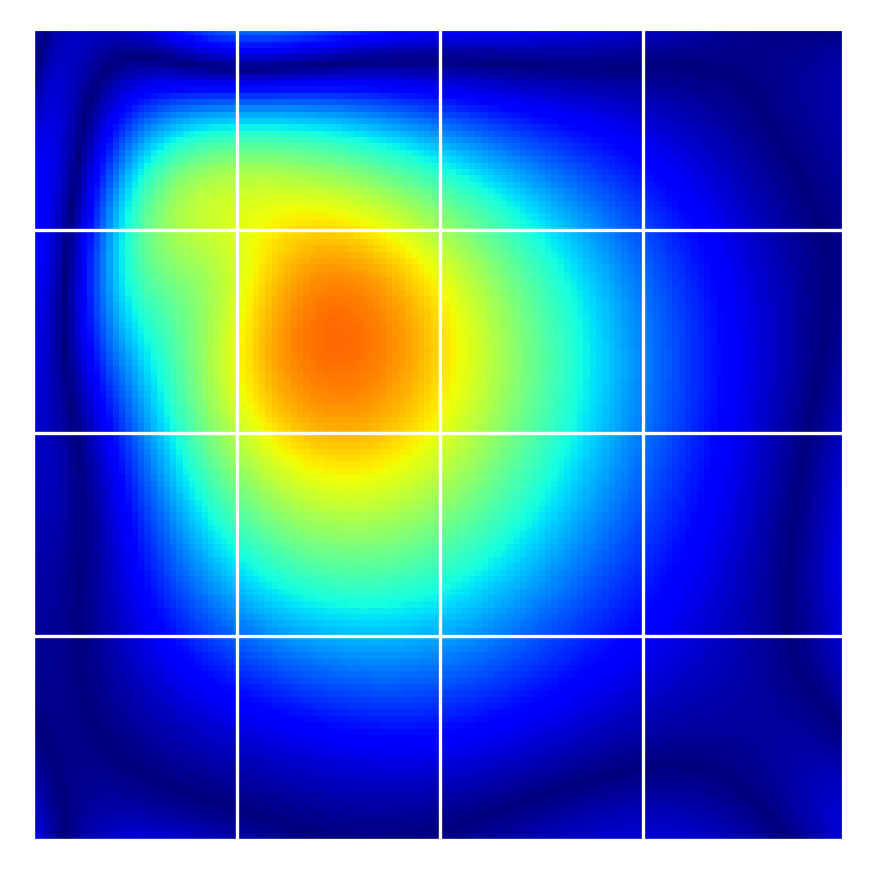}}%
\subcaptionbox{}{\includegraphics[height=.184\textwidth]{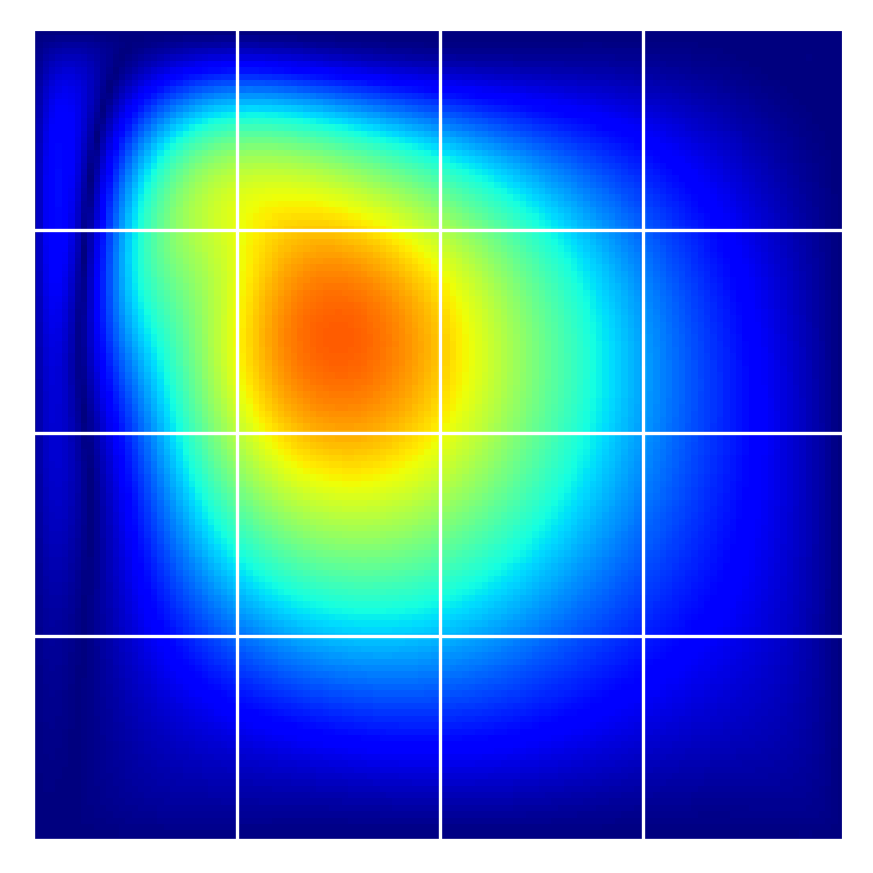}}%
\subcaptionbox{}{\includegraphics[height=.184\textwidth]{figs/colorbar/colorbar_004-0.png}}

\subcaptionbox{Ground Truth}{\includegraphics[height=.184\textwidth]{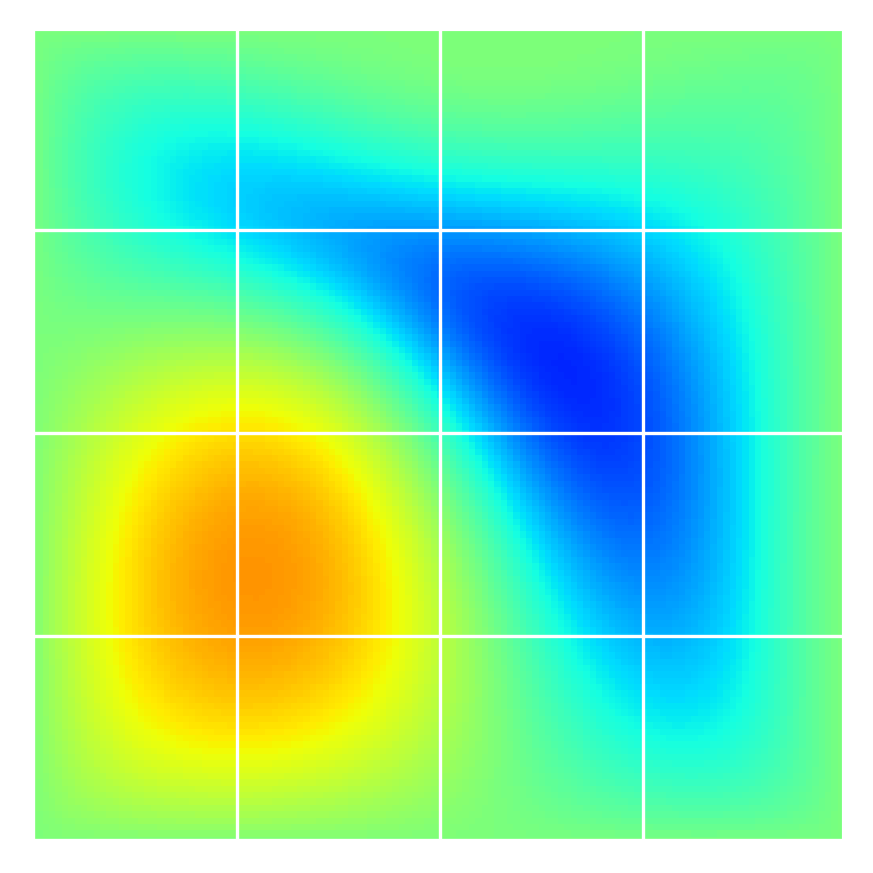}}%
\subcaptionbox{Prediction Soft BC}{\includegraphics[height=.184\textwidth]{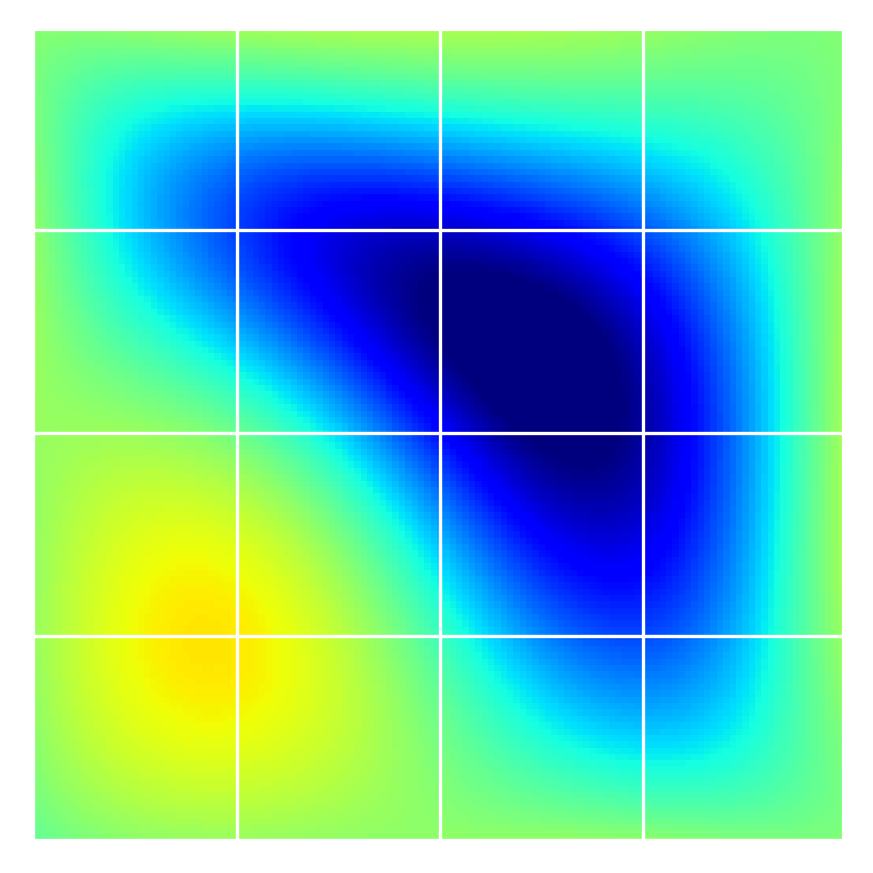}}%
\subcaptionbox{Prediction Exact BC}{\includegraphics[height=.184\textwidth]{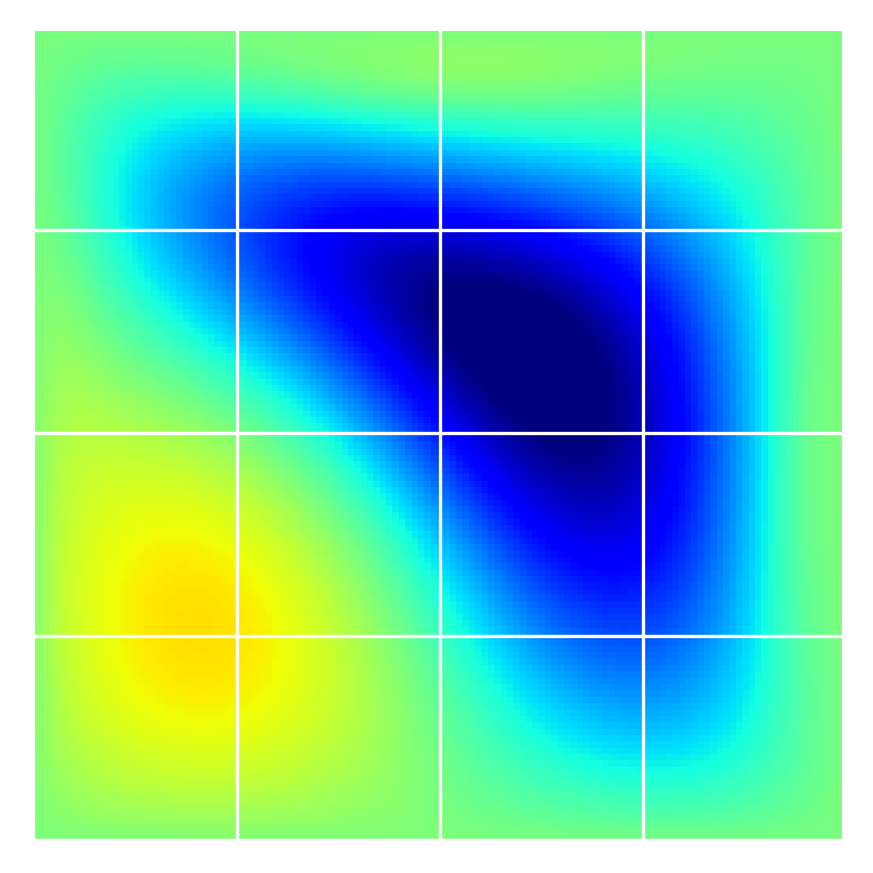}}%
\subcaptionbox{}{\includegraphics[height=.184\textwidth]{figs/colorbar/colorbar_05-05.png}}%
\subcaptionbox{Error Soft BC}{\includegraphics[height=.184\textwidth]{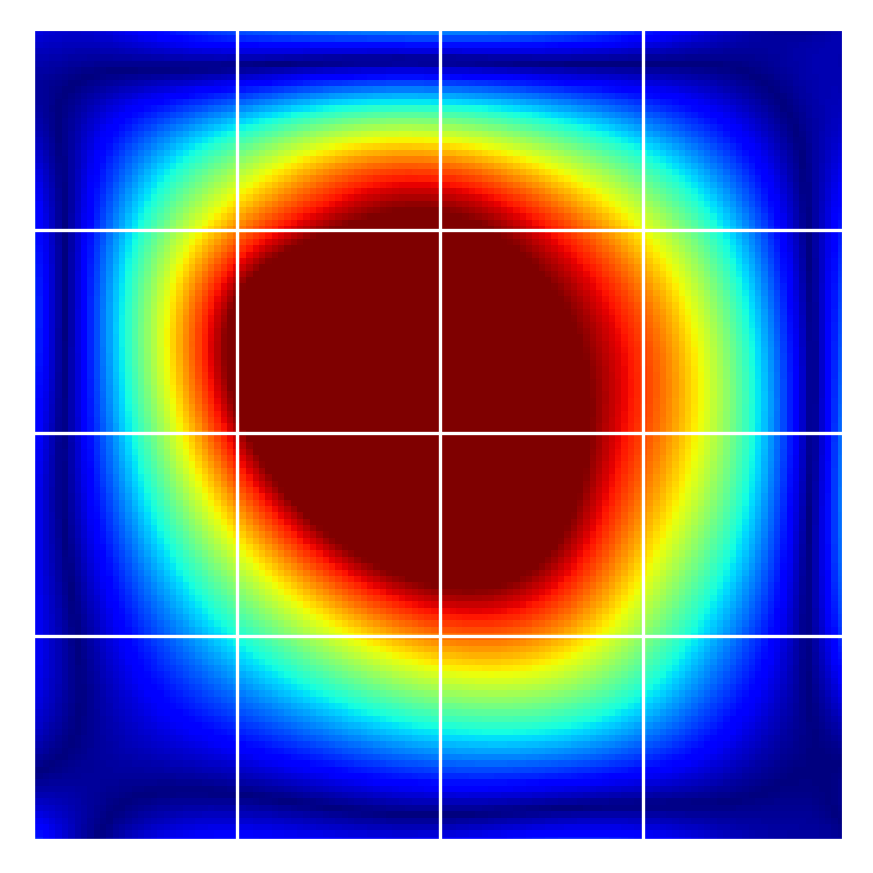}}%
\subcaptionbox{Error Exact BC}{\includegraphics[height=.184\textwidth]{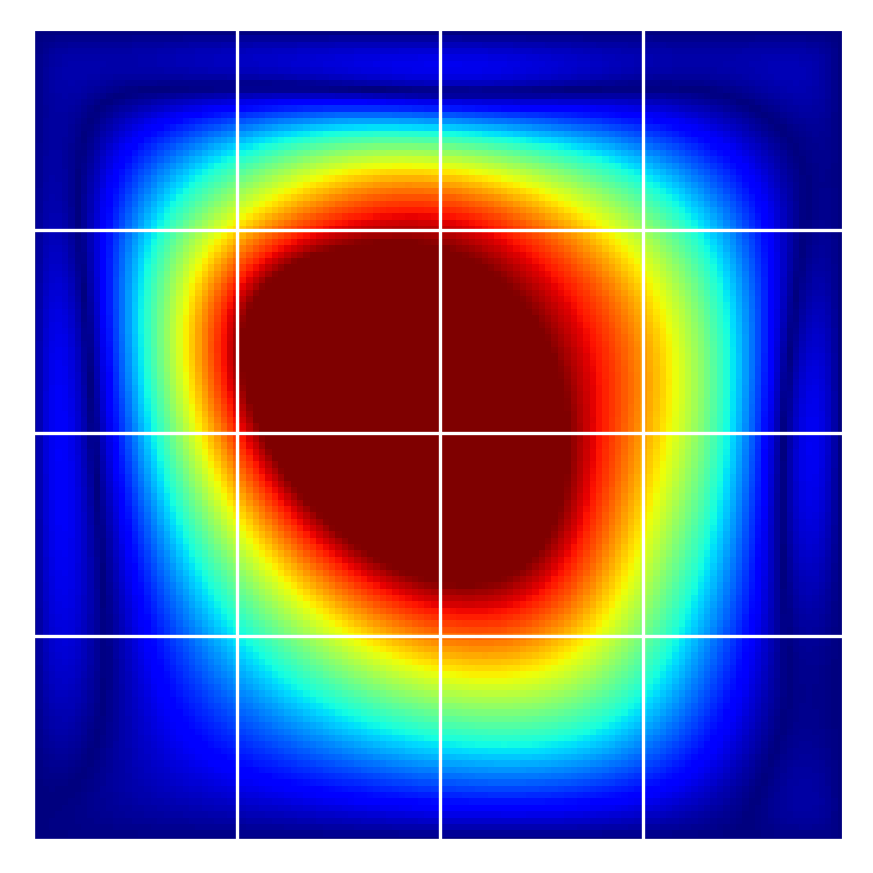}}%
\subcaptionbox{}{\includegraphics[height=.184\textwidth]{figs/colorbar/colorbar_02-0.png}}

\caption{Results from models solving Poisson's equation with zero BC (one dataset per row).}
\label{fig:models_eq0}
\end{figure}

\begin{table}[ht]
  \centering
  \begin{tabular}{clcccc}
    \toprule
    Dataset \#         & BC Type & MAE     & RMSE    & MAPE      & Avg. time / epoch (sec)\\
    \midrule
    \multirow{2}{*}{0} & Soft    & 6.08e-2 & 8.39e-2 & 152.46\%  & 1.77         \\
                       & Exact   & 5.79e-2 & 8.27e-2 & 119.99\%  & 1.86         \\
    \midrule
    \multirow{2}{*}{1} & Soft    & 1.23e-1 & 1.73e-1 & 397.30\%  & 1.78         \\
                       & Exact   & 1.06e-1 & 1.54e-1 & 262.48\%  & 1.84         \\
    \midrule
    \multirow{2}{*}{2} & Soft    & 9.10e-3 & 1.27e-2 & 433.58\% & 1.77          \\
                       & Exact   & 9.60e-3 & 1.31e-2 & 262.11\% & 1.81          \\
    \midrule
    \multirow{2}{*}{3} & Soft    & 9.72e-2 & 1.29e-1 & 375.03\% & 1.78          \\
                       & Exact   & 8.61e-2 & 1.19e-1 & 263.66\% & 1.83          \\
    \bottomrule\\
  \end{tabular}
  \caption{Error metrics for models solving Poisson's equation with zero BC (plots in Figure~\ref{fig:models_eq0})}
  \label{tab:models_eq0}
\end{table}

%%%%%%%%%%%%%%%%%%%%%%%%%%%%%%%%%%%%%%%%%%%%%%%%%%%%%%%%%%%%

\end{document}